\documentclass{article}
\usepackage[square]{wsnatbib}
\usepackage{amsmath}
\usepackage{algorithm}
\usepackage{algpseudocode}
\usepackage{multirow}
\usepackage[normalem]{ulem}
\useunder{\uline}{\ul}{}
\usepackage{subcaption}
\usepackage{graphicx}

\newcommand{\figref}[1]{\textbf{Fig.}~\ref{#1}}
\newcommand{\tabref}[1]{{\textbf{Table}~\ref{#1}}}
\renewcommand{\algref}[1]{\textbf{Algorithm~\ref{#1}}}

\begin{document}

%
%

\markboth{Y. He and C. Aranha}
{Solving Portfolio Optimization Problems Using MOEA/D and L\'evy Flight}

\title{Solving Portfolio Optimization Problems Using\\
MOEA/D and L\'evy Flight}

\author{Yifan He, Claus Aranha\\
he.yifan.xs@alumni.tsukuba.ac.jp, caranha@cs.tsukuba.ac.jp\\
The University of Tsukuba}




\maketitle


\begin{abstract}
Portfolio optimization is a financial task which requires the allocation of capital on a set of financial assets to achieve a better trade-off between return and risk. To solve this problem, recent studies applied multi-objective evolutionary algorithms (MOEAs) for its natural bi-objective structure. This paper presents a method injecting a distribution-based mutation method named L\'evy Flight into a decomposition based MOEA named MOEA/D. The proposed algorithm is compared with three MOEA/D-like algorithms, NSGA-II, and other distribution-based mutation methods on five portfolio optimization benchmarks sized from 31 to 225 in OR library without constraints, assessing with six metrics. Numerical results and statistical test indicate that this method can outperform comparison methods in most cases. We analyze how L\'evy Flight contributes to this improvement by promoting global search early in the optimization. We explain this improvement by considering the interaction between mutation method and the property of the problem.
\end{abstract}


\section{Introduction} \label{sec:introduction}
Multi-Objective Optimization Problems (MOOP) consist of several conflicting objectives and require that an optimization algorithm finds an optimal set of trade-offs rather than a single optimal solution. In the financial world, researchers and investors face a famous MOOP known as portfolio optimization (PO). The goal of this problem is to find an optimal allocation of capital among a finite set of available financial assets, by maximizing portfolio return and minimizing portfolio risk simultaneously. One of the challenges of this problem comes from its complex and large search space. Although the unconstrained model of modern portfolio theory contributed by Markowitz~\cite{Markowitz1952} states that PO can be solved using a quadratic programming method, realistic constraints make the problem NP-hard~\cite{Bienstock1996}.

\begin{figure}
    \centerline{\includegraphics[width=0.7\textwidth]{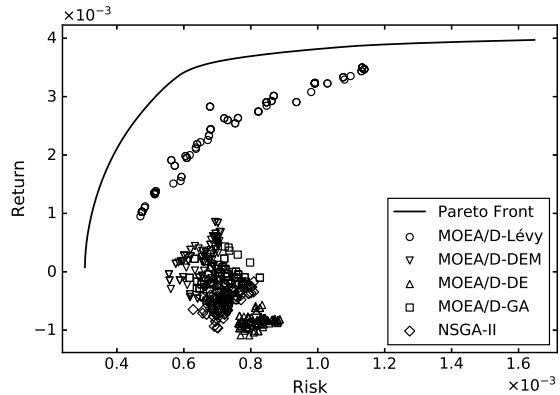}}
    \vspace*{8pt}
    \caption{Population of different methods for the Nikkei dataset at the 10th generation. The proposed method (MOEA/D-L\'evy) quickly explores a larger area of the objective space.} \label{fig:5alg10thpoponnikkei}
\end{figure}

To solve this complex problem, evolutionary algorithms (EAs) are frequently applied. By assigning different weights to each objective, the PO can be addressed in a single-objective form. However, considering its natural bi-objective formulation, researchers have paid increasing efforts on methods using multi-objective evolutionary algorithms (MOEAs). Some domination-based MOEAs, such as MOGA, PAES, SPEA2 and NSGA-II, have been assessed on PO benchmarks~\cite{Skolpadungket2007,Mishra2009a,Mishra2011,Anagnostopoulos2011a}, or applied on practical PO applications~\cite{Duran2009,Anagnostopoulos2011}. Also, several multi-objective variants of swarm intelligence methods, namely non-dominated sorting Multi-Objective Particle Swarm Optimization (NS-MOPSO)~\cite{Mishra2014}, Multi-Objective Bacteria Foraging Optimization (MOBFO)~\cite{Mishra2014a}, and Multi-objective Co-variance based Artificial Bee Colony (M-CABC)~\cite{Kumar2017}, have been introduced to the PO literature in the recent years. However, only a few PO researchers have paid attention to MOEA based on Decomposition (MOEA/D)~\cite{Zhang2010,Zhang2018,Zhou2018}. This powerful decomposition-based MOEA has not been well-discussed on PO yet.

Recently, researchers have paid attention to L\'evy Flight (LF) for solving hard optimization problems. Prior studies have shown the strong search capability of this mutation method~\cite{Viswanathan2008,Hakl2014}. In this paper, we propose a method named MOEA/D-L\'evy, which injects LF into MOEA/D, and assess it on PO with unit constraint (\textbf{Section~\ref{sec:proposed-method}}). This modification is motivated by the efficient global search performance of LF. As one of the main challenges in the PO is the high-dimensionality of the search space, we expect the global search capability of LF can overcome this difficulty during optimization. Our experiments (\textbf{Section~\ref{sec:experiments}}) include a comparison with literature methods and a comparison with mutations based on different probability distributions. The results of six evaluation metrics, as well as a statistical test on all five datasets in a frequently used PO benchmark in OR Library~\cite{Chang2000}, indicate that this method outperforms the comparison methods in most cases. Additionally, we use an experiment to show how the addition of LF contributes to the optimization process (\textbf{Section 5}). \figref{fig:5alg10thpoponnikkei} illustrates this contribution, by showing the population of five methods on the objective space at the 10-th generation when optimizing for the Nikkei dataset. It is interesting to notice that at the very beginning of optimization the proposed method, represented by circles, can achieve a relatively good solution set compared to other literature methods. We suggest this may be caused by a compound factor of LF and the repair method, as well as the characteristics of the PO problem. To our best knowledge, no research has applied LF into MOEA/D in the PO literature. The code and data used for the experiments in this paper are available at a public repository\footnote{https://github.com/Y1fanHE/po\_with\_moead-levy}.

\section{Background} \label{sec:background}
\subsection{Portfolio Optimization} \label{sec:portfolio-optimization}
\begin{figure}
    \centerline{\includegraphics[width=0.7\textwidth]{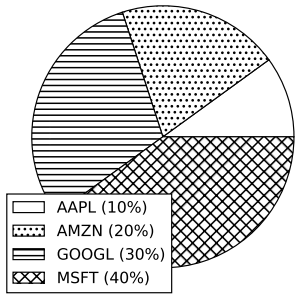}}
    \caption{A sample portfolio on four assets} \label{fig:porteg}
\end{figure}

PO requires an allocation of capital among a set of available financial assets to achieve a better trade-off between return and risk. \figref{fig:porteg} provides an example of a portfolio on four stocks. The whole pie represents all capital  and the rate in the sectors represents invest rate on the corresponding assets (e.g. 30\% of the capital has been allocated into Google stock). The initial work by Markowitz~\cite{Markowitz1952} models PO as the following bi-objective formulation. In this PO model, the first objective (return) is being maximized, while the second objective (risk) is being minimized. This model requires to allocate all of the capital during the investment. This is usually represented as the sum of investments being equal to one (unit constraint). $\boldsymbol{w}=(w_1, w_2, ..., w_n)$ is a vector representing the invest ratios of $n$ assets. $r_i$ is the return rate of $i$-th asset, and $\sigma_{ij}$ is co-variance between the return rated of $i$-th asset and $j$-th asset.

\begin{equation} \label{eq:m-vreturn}
    \max_{\boldsymbol{w}} RETURN = \sum_{i=1}^{n} r_i w_i
\end{equation}
\begin{equation} \label{eq:m-vrisk}
    \min_{\boldsymbol{w}} RISK = \sum_{i=1}^{n} \sum_{j=1}^{n} \sigma_{ij} w_i w_j
\end{equation}
subject to,
\begin{equation} \label{eq:m-vconstraints}
    \sum_{i=1}^{n} w_i = 1, (0 \leq w_i \leq 1)
\end{equation}

It is easy to notice that PO belongs to the big category of multi-objective optimization problems (MOOPs). MOOP is a class of optimization problems which contain more than one conflicting objectives. Compared to single-objective optimization problems (SOOPs), MOOPs aim  to  achieve  an optimal set  of solution showing best trade-off between its multiple objectives rather than one single optimal solution. Such an optimal set named Pareto Front contains all the feasible solutions that are not dominated by any other solutions in problem feasible region.

\subsection{Multi-Objective Evolutionary Algorithms} \label{sec:multi-objective-evolutionary-algorithm}
There is a large number of MOEAs. Many MOEAs, such as NSGA-II~\cite{Deb2002}, find a set of trade-off solutions using the concept of \emph{domination}, where two solutions are \emph{non-dominates} if each solution has at least one objective where it is better than the other one. As an alternative approach, MOEA/D~\cite{QingfuZhang2007} achieves Pareto Front by decomposing a MOOP into several single objective optimization problems (SOOP), and optimizing them simultaneously. One of the most frequently used decomposition methods is Tchebycheff approach. This approach transfers vector optimization into scalar optimization with a weight vector $\boldsymbol{\lambda}$ and a reference point $\boldsymbol{z^*}$ as follows (i.e.,minimization example), where $z^*_m$ is the minimum value of $m$-th objective. By using this method, the retrieved optimal is the intersection of Pareto Front and the straight line determined by $\boldsymbol{\lambda}$ and $\boldsymbol{z^*}$.

\begin{equation} \label{eq:tch}
    \min_{\boldsymbol{x}} g^{te}(\boldsymbol{x} \mid \boldsymbol{\lambda}, \boldsymbol{z^*}) =
    \max_{m=1,...,M} \{\lambda_m |f_m(\boldsymbol{x}) - z^*_m|\}
\end{equation}
\begin{equation} \label{eq:tchlambda}
    \boldsymbol{\lambda}=(\lambda_1,...,\lambda_M), \lambda_1+...+\lambda_M=1
\end{equation}
\begin{equation} \label{eq:tchz*}
    \boldsymbol{z^*}=(\min f_1,...,\min f_M)
\end{equation}

In MOEA/D, every individual in the population is optimized as the solution of one SOOP decomposed using one weight vector. During optimization, the solution uses the information of its neighbor. The neighbor is defined as solutions of neighboring SOOPs, which can be computed using the closest Euclidean distance between weight vectors. When reproducing an offspring, MOEA/D selects parents from the neighbors of the current individual. Once the offspring are generated, it can be used to update all neighbors of the current individual. \algref{alg:moead} shows a detailed description of MOEA/D.

\begin{algorithm}[t]
    \caption{Original MOEA/D}
    \label{alg:moead}
    \begin{algorithmic}[1]
        \State Generate a set of weight vectors $\{\boldsymbol{\lambda^1},...,\boldsymbol{\lambda^N}\}$;
        \State Determine neighbor;
        \State Initialize population $\{\boldsymbol{x^1},...,\boldsymbol{x^N}\}$;
        \State Compute reference point $\boldsymbol{z^*}$;
        \While {stopping criteria}
            \For {$\boldsymbol{x^i}$ in population}
                \State Select parents from neighbor of $\boldsymbol{x^i}$;
                \State Reproduce an offspring $\boldsymbol{y}$ by GA operator;
                \State Update reference point $\boldsymbol{z^*}$;
                \For {$\boldsymbol{x^p}$ in neighbor of $\boldsymbol{x^i}$}
                    \If {$g^{te}(\boldsymbol{y}\mid\boldsymbol{\lambda^p},\boldsymbol{z^*})\leq g^{te}(\boldsymbol{x^p}\mid\boldsymbol{\lambda^p},\boldsymbol{z^*})$}
                        \State Set $\boldsymbol{x^p}=\boldsymbol{y}$;
                    \EndIf
                \EndFor
            \EndFor
        \EndWhile
    \end{algorithmic}
\end{algorithm}

Recently, researchers have tried to combine MOEA/D with mutation methods in other meta-heuristics to enhance its search capability, such as Particle Swarm Optimization (PSO), Differential Evolution (DE), and Ant Colony Optimization (ACO)~\cite{WeiPeng2008,HuiLi2009,Ke2013}. MOEA/D-DE~\cite{HuiLi2009} injects the DE operator and polynomial mutation operator into MOEA/D. The DE operator is present as follows, where $\boldsymbol{x^i}$, $\boldsymbol{x^j}$ and $\boldsymbol{x^k}$ are parents, $\boldsymbol{y}$ is offspring, and $rand$ is a random number between 0 and 1.

\begin{equation} \label{eq:de}
    \boldsymbol{y}=
    \begin{cases}
        \boldsymbol{x^i}+F\cdot(\boldsymbol{x^j}-\boldsymbol{x^k}), & rand<CR\\
        \boldsymbol{x^i}, & rand\geq CR
    \end{cases}
\end{equation}

The original DE operator includes a DE mutation step and a crossover step with the original parent. However, the authors of MOEA/D-DE have suggested setting $CR$ to 1.0 to deal with complicated problems, which means only the DE mutation step will be implemented. Additionally, they have designed a diversity keeping strategy, including an upper limitation for updating neighbor, and a small probability to select parents from the whole population rather than the neighbor. The parent selection scheme of MOEA/D-DE has been well discussed by Tanabe~\cite{Tanabe2019}. This study has reported that using curr/1 (i.e., select current individual as $\boldsymbol{x^i}$) and WR (i.e., $\boldsymbol{x^i}$, $\boldsymbol{x^j}$, $\boldsymbol{x^k}$ can be the same individual) or WPR (i.e., $\boldsymbol{x^j}$ and $\boldsymbol{x^k}$ cannot be same, but either can be the same as $\boldsymbol{x^i}$) outperforms other settings.

\subsection{Portfolio Optimization Using MOEAs} \label{sec:portfolio-optimization-using-moeas}
Since the genetic algorithm was firstly introduced into PO literature in a multi-objective form in 1993~\cite{Arnone1993}, prior studies have assessed a large group of MOEAs. One frequently used benchmark in these assessments is the OR library~\cite{Chang2000}, which contains five PO datasets, namely Hangseng, DAX 100, FTSE 100, S\&P 100 and Nikkei. VEGA, Fuzzy VEGA, MOGA, SPEA2, and NSGA-II have been compared on Hangseng with constraints~\cite{Skolpadungket2007}. Another two comparison studies using the same dataset have reported that NSGA-II outperforms PESA, PAES and APAES~\cite{Mishra2009a,Mishra2011}. NSGA-II, PESA, and SPEA2 have also been assessed on DAX 100 with constraints~\cite{Anagnostopoulos2011a}. The results have shown that NSGA-II and SPEA2 hold the best average performance. In Branke’s study~\cite{Branke2009}, envelope-based MOEA has been assessed on Hangseng, S\&P 100, and Nikkei with realistic constraints.

While EAs are modeled from the evolutionary process, swarm intelligence (SI) is a group of algorithms based on the self-organization of individuals. A survey has reported increasing attention on using MOEAs with SI to solve PO~\cite{Ertenlice2018}. MOPSO has been compared with PSFGA, SPEA2, and NSGA-II on Hangseng~\cite{Mishra2009}. The results have indicated that MOPSO outperforms the other three methods significantly. NS-MOPSO, MOBFO, and M-CABC have been proposed and assessed on all five PO datasets in the OR library with constraints~\cite{Mishra2014,Mishra2014a,Kumar2017}.

Despite performance assessment using benchmarks, researchers have also shown strong interest in developing MOEAs for practical PO. NSGA-II, SPEA2, and IBEA have been compared on solving PO with financial data in the Venezuelan market~\cite{Duran2009}. Five domination based MOEAs have been tested on cardinality constrained PO with a dataset containing over 2000 assets~\cite{Anagnostopoulos2011}. Variants of MOPSO have been proposed to solve the constrained PO with realistic dataset~\cite{Liang2013,JianliZhou2014}. Several studies have focused on designing specific initialization methods, problem guided mutation and constraint handling techniques on PO~\cite{Orito2013,Liagkouras2014,Meghwani2018,Liagkouras2018}.

While most of the prior studies are based on domination methods, few researchers have focused on solving PO with decomposition-based MOEAs. Zhang’s study~\cite{Zhang2010} has proposed a new decomposition method and used MOEA/D-DE to solve a constrained PO. The experimental results show that MOEA/D-DE performs better than NSGA-II-DE. A new weight vector generation approach to achieve an evenly distributed vector set has been proposed by Zhang~\cite{Zhang2018}. In Zhou’s study~\cite{Zhou2018}, researchers have combined data envelopment analysis techniques with MOEA/D and assessed this method on ZDT1-3 benchmarks as well as PO application on 10 stocks. The results show that the proposed method can outperform MOEA/D. Thus, performance assessment and application of MOEA/D on PO have not been well discussed so far.

In Zhang’s study~\cite{Zhang2010}, researchers have reported that original Tchebycheff decomposition cannot achieve an evenly distributed optimal set, for the scale of two objectives are usually different in PO. To solve this problem, they have proposed the NBI-style Tchebycheff decomposition approach (i.e., NBI: Normal Boundary Intersection). This approach transfers vector optimization into scalar optimizations with a normal vector $\boldsymbol{\lambda}$ of convex hull of minima (CHIM) and evenly distributed reference points $\{\boldsymbol{r^1},...,\boldsymbol{r^N}\}$ on CHIM as follows (i.e., bi-objective minimization example), where $\boldsymbol{F^1}$ and $\boldsymbol{F^2}$ are extreme points. By using this method, the $i$-th optimal is the intersection of Pareto Front and the normal line of determined by $\boldsymbol{\lambda}$ and $\boldsymbol{r^i}$. In the case of bi-objective optimization problems, CHIM is the straight line between extreme points.

\begin{equation} \label{eq:nbi}
    \min_{\boldsymbol{x}} g^{tn}(\boldsymbol{x} \mid \boldsymbol{\lambda}, \boldsymbol{r}) =
    \max_{m=1,2} \{\lambda_m (f_m(\boldsymbol{x}) - r_m)\}
\end{equation}
\begin{equation} \label{eq:nbilambda}
    \boldsymbol{\lambda}=(\lambda_1,\lambda_2), \lambda_1=|F^2_2-F^1_2|, \lambda_2=|F^2_1-F^1_1|
\end{equation}
\begin{equation} \label{eq:nbir}
    \boldsymbol{r^i}=a_i\cdot\boldsymbol{F^1} + (1-a_i)\cdot\boldsymbol{F^2}, a_i=\frac{N-i}{N-i}
\end{equation}

To generate an offspring, they have used DE mutation operator. In that work, the three parents are randomly selected from neighbor of current individual. What is more, although original MOEA/D-DE~\cite{HuiLi2009} uses diversity keeping strategies, the usage of such strategies has not been reported in Zhang's study.

\subsection{L\'evy Flight and Optimization} \label{sec:levy-flight-and-optimization}
\begin{figure}
    \centerline{\includegraphics[width=0.7\textwidth]{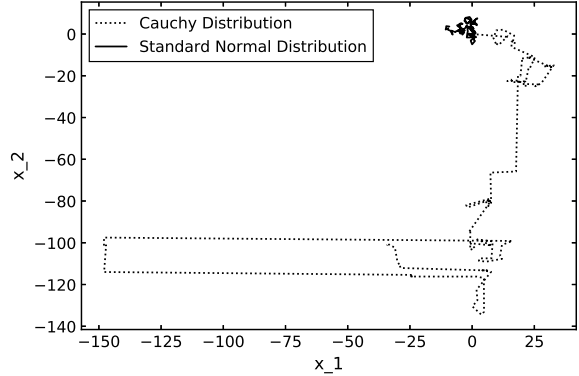}}
    \vspace*{8pt}
    \caption{A comparison on random walks using the normal distribution and the Cauchy distribution (L\'evy Flight). Using occasional long jumps, LF explores a larger area than the Normal random walk.}\label{fig:levyeg}
\end{figure}
L\'evy Flight (LF) is a special random walk where step length is drawn from a L\'evy-stable distribution (a heavy-tailed distribution). Thus, LF is a process mixing short motions and long trajectories. This property helps escape from local optimal, and thus improves the performance in an optimization task. \figref{fig:levyeg} shows a comparison of a 100-step random walk in two-dimensional space using a standard normal distribution and Cauchy distribution (i.e., an example of heavy-tail distribution). The search area of the random walk using Cauchy distribution is much larger than the random walk using the standard normal distribution. One biological application of LF is the LF foraging theory~\cite{Viswanathan2008}. This theory states that creatures have evolved to use LF during foraging for its optimized random search capability. Despite this natural example, LF has been applied in optimization problems in various areas such as physics, biology, statistics, finance, and economics~\cite{Kamaruzaman2013}.

In addition, researchers have developed and enhanced metaheuristics using LF. Cuckoo Search (CS)~\cite{Yang2009} is an efficient optimization method using LF to implement a global search. Researchers have also used LF to enhance PSO and ABC~\cite{Hakl2014,Ma2015,Jensi2016,Aydogdu2016}. In Zhang's study~\cite{Zhang2019}, a modified CS has been injected into MOEA/D to solve a spectrum allocation problem. However, the mis-setting of evaluation times (i.e., they do not set an equal evaluation time for all algorithms in the experiment) and the unclear algorithm description (i.e., they do not report some details in the numerical process when describing algorithm procedure) may confuse other researchers when understanding and implementing this method. Thus, there remains a proper assessment of MOEA/D injected with LF. Some studies have applied the single-objective form of CS to solve PO~\cite{Shadkam2015,FaezyRazi2016,El-Bizri2017}. However, to our best knowledge, there is no research injecting LF into MOEA/D in the PO literature.

Cauchy distribution is one of the stable distributions. Thus, Cauchy mutation can be seen as a special case of LF. Prior studies have shown the superiority of Cauchy mutation in diversity keeping. Ali has proposed a hybrid method based on DE and Cauchy mutation.~\cite{Ali2011} Ali's algorithm performs Cauchy mutation on the best individual when DE mutation fails to update in several continuous generations. This modification is proved to be efficient to avoid the premature issue of DE in their experiments. However, this algorithm does not take Cauchy mutation as the main method. Several works have designed adaptive strategies for DE based on Cauchy distribution~\cite{Choi2013,Zhang2009}. These algorithms generate parameters from the Cauchy distribution and update the expectation value of Cauchy distribution based on successfully updates solutions.

\section{Proposed Method} \label{sec:proposed-method}
\subsection{L\'evy Flight Mutation} \label{sec:levy-flight-mutation}
The LF mutation in the proposed method is similar to the DE mutation, utilizing the difference between individuals. However, the scaling factor in LF mutation is a vector generated from heavy-tail distribution rather than a constant. What is more, LF mutation only uses two parents, while DE mutation uses three parents. The formulation of this mutation method is as follows, where $\boldsymbol{x^i}$ and $\boldsymbol{x^j}$ are parents, $\boldsymbol{y}$ is offspring, $\oplus$ means entry-wise multiplication, $\alpha_0$ is a scaling factor, and $\boldsymbol{L\acute{e}vy(\beta)}$ is a vector where each component is generated using Mantegna’s algorithm (MA)~\cite{Mantegna1994} (i.e., $0.3 \leq \beta \leq 1.99$). This algorithm generates random numbers from symmetric L\'evy-stable distribution. MA is present following LF mutation, where $\Gamma$ represents the Gamma function.

\begin{equation} \label{eq:lfmutation}
    \boldsymbol{y}=\boldsymbol{x^i}+\alpha_0\cdot(\boldsymbol{x^i}-\boldsymbol{x^j})\oplus\boldsymbol{L\acute{e}vy(\beta)}
\end{equation}
\begin{equation} \label{eq:ma}
    \boldsymbol{L\acute{e}vy(\beta)}\sim\frac{\boldsymbol{u}}{\boldsymbol{|v|}^{1/\beta}},
    \boldsymbol{u}\sim \boldsymbol{N}(0,\sigma_u^2), \boldsymbol{v}\sim \boldsymbol{N}(0,\sigma_v^2)
\end{equation}
\begin{equation} \label{eq:masigma}
    \sigma_u = \left\{ \frac{\Gamma(1+\beta)\sin(\pi\beta/2)}{\Gamma[(1+\beta)/2]\beta 2^{(\beta-1)/2}}\right\}^{1/\beta},
    \sigma_v = 1
\end{equation}

$\beta$ is the stability parameter controlling the shape of a stable distribution, and thus can control the balance between local and global search in LF. A smaller $\beta$ holds a stronger global search capability. When $\beta$ is set to 1, MA generates Cauchy random numbers. However, there are several differences between our LF mutation and Cauchy mutation that mentioned in the previous section. First, compared with Ali's work~\cite{Ali2011}, our method applies LF mutation in all generations. Second, compared with adaptive strategies based on Cauchy distribution~\cite{Choi2013,Zhang2009}, our mutation method performs entry-wise multiplication, while the adaptive strategy only generates constant scaling factors, and our method is not an adaptive strategy. Third, in our later experiment, we use $\beta=0.3$ rather than 1. Thus, we actually are not using Cauchy mutation but a more general method. What is more, this mutation method not only does a diversification but also utilizes the property of the PO problem (we show this in the later experiments).

\subsection{Repair Method} \label{sec:repair-method}
The offspring generated by LF mutation may not satisfy the unit constraint in~\eqref{eq:m-vconstraints}. Thus, it is necessary to apply a proper constraint handling technique after reproduction. In this research, a general repair method following the description in~\algref{alg:repair} has been applied. These repair steps will first set negative variables to 0, and then scale the entire vector (offspring) so that the summation of all variables equals to 1.

\begin{algorithm}[t]
    \caption{Repair Method}
    \label{alg:repair}
    \begin{algorithmic}[1]
        \For{$y_i$ in offspring $\boldsymbol{y}$}
            \If{$y_i < 0$}
                \State Set $y_i = 0$;
            \EndIf
        \EndFor
        \State Compute $s = \sum y_i$;
        \For{$y_i$ in offspring $\boldsymbol{y}$}
            \State Set $y_i = y_i / s$;
        \EndFor
    \end{algorithmic}
\end{algorithm}

\begin{algorithm}[t]
    \caption{Proposed MOEA/D-L\'evy}
    \label{alg:moead-levy}
    \begin{algorithmic}[1]
        \State Determine neighbor;
        \State Initialize population $\{\boldsymbol{x^1},...,\boldsymbol{x^N}\}$;
        \State Compute extreme points $\boldsymbol{F^1},\boldsymbol{F^2}$;
        \While {stopping criteria}
            \For {$\boldsymbol{x^i}$ in population}
                \If{$rand(0,1)<\sigma$}
                    \State Set $B(i)$ as neighbor of $\boldsymbol{x^i}$;
                \Else
                    \State Set $B(i)$ as population $\{\boldsymbol{x^1},...,\boldsymbol{x^N}\}$;
                \EndIf
                \State Select parents from $B(i)$;
                \State Reproduce an offspring $\boldsymbol{y}$ by LF mutation and polynomial mutation;
                \State Repair $\boldsymbol{y}$ to satisfy constraints;
                \State Update extreme points $\boldsymbol{F^1},\boldsymbol{F^2}$;
                \State Set update counter $n_c=0$;
                \State Re-arange $B(i)$ in random order;
                \For {$\boldsymbol{x^p}$ in neighbor of $B(i)$}
                    \If {$g^{tn}(\boldsymbol{y}\mid\boldsymbol{\lambda},\boldsymbol{r^p})\leq g^{tn}(\boldsymbol{x^p}\mid\boldsymbol{\lambda},\boldsymbol{r^p})$}
                        \State Set $\boldsymbol{x^p}=\boldsymbol{y}$;
                        \State Set $n_c = n_c + 1$;
                        \If{$n_c \geq n_r$}
                            \State Break;
                        \EndIf
                    \EndIf
                \EndFor
            \EndFor
        \EndWhile
    \end{algorithmic}
\end{algorithm}

\subsection{Proposed MOEA/D-L\'evy Algorithm} \label{sec:proposed-moead-levy-algorithm}
An entire algorithmic description of the proposed method is presented in~\algref{alg:moead-levy}. In the following sections, the proposed method will be named as MOEA/D-L\'evy for convenience. As our problem only includes unit constraint, the representation of a portfolio in our algorithm is a vector, where each component is an invest rate of one asset. In MOEA/D-L\'evy, each variable of a solution is initialized using a uniform distribution between 0 and 1. When selecting parents, the current individual is selected as $\boldsymbol{x^i}$, while $\boldsymbol{x^j}$ is randomly selected from the neighbors of the current individual. MOEA/D-L\'evy injects LF mutation and polynomial mutation operators into MOEA/D. When an offspring is generated, the algorithm will implement repair steps to satisfy constraints in~\eqref{eq:m-vconstraints}. NBI-style Tchebycheff decomposition approach (i.e., described in \textbf{Section~\ref{sec:portfolio-optimization-using-moeas}}) has been applied to deal with different scales of two objectives in PO. We have also applied the diversity keeping strategies proposed in Li’s study~\cite{HuiLi2009} (i.e., described in \textbf{Section~\ref{sec:multi-objective-evolutionary-algorithm}}), including a small proportion $1-\sigma$ to select parents from the whole population (i.e.,  described in~\algref{alg:moead-levy}, Line 6 to 10), where $rand(0,1)$ means a random number generated from uniform distribution with range 0 to 1, as well as an upper limitation $n_r$ for updating neighbor (i.e., described in~\algref{alg:moead-levy}, Line 20 to 23). The normal vector $\boldsymbol{\lambda}$ and reference point $\boldsymbol{r^p}$ are computed based on ~\eqref{eq:nbilambda} and~\eqref{eq:nbir}.

Compared with the original MOEA/D algorithm~\cite{QingfuZhang2007}, the main modification of MOEA/D-L\'evy includes NBI-style Tchebycheff decomposition (the original MOEA/D applies Tchebycheff method), diversity keeping strategies (the original MOEA/D does not apply it), and reproduction method based on LF and polynomial mutation (the original MOEA/D applies GA operator).

\section{Experiments} \label{sec:experiments}
\subsection{Experimental Method}\label{sec:experimental-method}
In this study, the goal is to assess the performance of LF as a mutation method in MOEA/D. Therefore, in Experiment I we compare MOEA/D-L\'evy with three variants of MOEA/D and NSGA-II. In addition, to show the effectiveness of LF, in Experiment II we compare variants of the proposed algorithm using different distributions. The benchmarks used are all five PO datasets in the OR library~\cite{Chang2000}, containing 31, 85, 89, 98, and 225 assets from 1992 to 1997, respectively. A summary of the datasets is presented in Table 1.

When evaluating MOEAs, it is important to consider both convergence and diversity performance. Although there are several metrics assessing convergence, such as generation distance (GD), and diversity, such as spacing (S), maximum spread (MS), spread ($\Delta$), a recent trend in the literature is to assess the two performance at one time using overall metrics, such as inverted generation distance (IGD) and hypervolume (HV). In this study, all six metrics (i.e., GD, S, MS, $\Delta$, IGD and HV) are applied. The calculation methods of GD, S, MS, $\Delta$ and HV are referred to Chapter 8 in Deb’s book~\cite{Deb2001}. For IGD, we refer to Coello’s study~\cite{CoelloCoello2004} which first proposed the IGD metric. A smaller value in GD, S, $\Delta$ and IGD indicates better performance, while for MS and HV, a larger value represents better performance.

\begin{itemize}
    \item \textbf{GD}: $d(\boldsymbol{v},P^*)$ is the minimum Euclidean distance between a non-dominated solution $\boldsymbol{v}$ and Pareto Front $P^*$.
    \begin{equation} \label{eq:gd}
        \frac{\sum_{\boldsymbol{v}\in A} d(\boldsymbol{v},P^*)}{|A|}
    \end{equation}

    \item \textbf{S}: $d_i$ is the minimum \textbf{\textit{Manhattan distance}} between $i$-th non-dominated solution and another one.
    \begin{equation} \label{eq:s}
        \sqrt{{\frac{1}{N^{'}}} \sum_{i=1}^{N^{'}} (\bar{d}-d_i)^2}
    \end{equation}

    \item \textbf{MS}: $f_m^i$ is the $m$-th objective of $i$-th non-dominated solution.
    \begin{equation} \label{eq:ms}
        \sqrt{\sum_{m=1}^M (\max_{i=1,...,N^{'}}f_m^i - \min_{i=1,...,N^{'}}f_m^i)^2}
    \end{equation}

    \item $\boldsymbol{\Delta}$: $d_i$ is the Euclidean distance between two consecutive non-dominated solutions; $d_f$ and $d_l$ are the Euclidean distance between extreme solutions of Pareto Front and boundary solutions of non-dominated set.
    \begin{equation} \label{eq:delta}
        \frac{d_f+d_l+\sum_{i=1}^{N^{'}-1} |d_i-\bar{d}|}{d_f+d_l+(N^{'}-1)\bar{d}}
    \end{equation}

    \item \textbf{IGD}: $d(\boldsymbol{v},A)$ is the minimum Euclidean distance between a solution $\boldsymbol{v}$ in Pareto Front $P^*$ and non-dominated set $A$.
    \begin{equation} \label{eq:igd}
        \frac{\sum_{\boldsymbol{v}\in P^*} d(\boldsymbol{v},A)}{|P^*|}
    \end{equation}

    \item \textbf{HV}: $v_i$ is the hypercube constructed with a reference point and $i$-th non-dominated solution as the diagonal corners. To compute the reference point, the Nadir point of solutions generated by all algorithms in the final generation will be used.
    \begin{equation} \label{eq:hv}
        \text{volume}\left( \bigcup_{i=1}^{N^{'}} v_i \right)
    \end{equation}
\end{itemize}

Both experiments are implemented with 51 repetitions. For every single run, the values of the six metrics at the final generation are calculated. All five datasets share the same parameter setting. The parameters are set as follows. For all methods, population size and maximum generation are set to 100 and 1500. An early stop criterion (convergence) is set when the variation of IGD is not larger than 1e-05 for 100 continuous generations. Neighbor size $T$, proportion $\sigma$ to select parents from the neighbors and upper limitation $n_r$ for updating neighbors in MOEA/D-based methods are set to 20, 0.9 and 2. These settings (i.e., $T$, $\sigma$ and $n_r$) are the same as Li’s study~\cite{HuiLi2009}.

\begin{figure}
    \centerline{\includegraphics[width=0.7\textwidth]{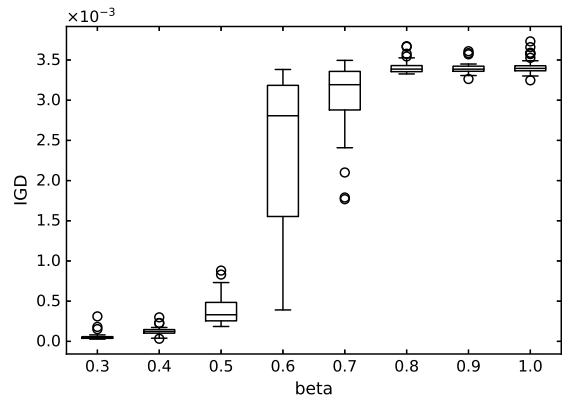}}
    \vspace*{8pt}
    \caption{An example of parameter tuning on MOEA/D-L\'evy} \label{fig:tuneg}
\end{figure}

The parameter settings of the mutation methods that differentiate the compared algorithms will be presented in \textbf{Section~\ref{sec:experiment-i-comparison-with-literature-methods}} and \textbf{Section~\ref{sec:experiment-ii-comparison-with-other-distribution-based-mutation-methods}}. These parameters are fine-tuned by a pre-experiment, using the Nikkei dataset. We perform 30 runs, stopping at the 300-th generation, and choose the parameters that receive the best average IGD to be used in the formal experiments. \figref{fig:tuneg} shows an example for $\beta$ parameter tuning on the proposed algorithm.

\begin{table}
\centering
\caption{A summary of the datasets used in experiments}
\label{tab:dataset}
\begin{tabular}{ccccc}
\hline
Dataset  & Region   & Size & Time Period                     \\ \hline
Hangseng & Hongkong & 31   & \multirow{5}{*}{1992$\sim$1997} \\
DAX 100  & Germany  & 85   &                                 \\
FTSE 100 & U.K.     & 89   &                                 \\
S\&P 100 & U.S.     & 98   &                                 \\
Nikkei   & Japan    & 225  &                                 \\ \hline
\end{tabular}
\end{table}

\begin{table}
\centering
\caption{A summary of mutation methods in Experiment I}
\label{tab:exp1}
\begin{tabular}{cc}
\hline
Method & Mutation Formula \\ \hline
\multirow{2}{*}{MOEA/D-L\'evy} & $\boldsymbol{y}=\boldsymbol{x^i}+\alpha_0 \cdot (\boldsymbol{x^i}-\boldsymbol{x^j}) \oplus \boldsymbol{L\acute{e}vy(\beta)}$ \\
 & polynomial mutation \\
\multirow{2}{*}{MOEA/D-DEM} & $\boldsymbol{y}=\boldsymbol{x^i}+F \cdot (\boldsymbol{x^j}-\boldsymbol{x^k})$ \\
 & polynomial mutation \\
MOEA/D-DE & $\boldsymbol{y}=\boldsymbol{x^i}+F \cdot (\boldsymbol{x^j}-\boldsymbol{x^k})$ \\
MOEA/D-GA & \multirow{2}{*}{\begin{tabular}[c]{@{}c@{}}SBX crossover and\\ polynomial mutation\end{tabular}} \\
NSGA-II &  \\ \hline
\end{tabular}
\end{table}

\begin{table}
\centering
\caption{A summary of mutation methods in Experiment II}
\label{tab:exp2}
\begin{tabular}{cc}
\hline
Method & Mutation Formula \\ \hline
LEVY & $\boldsymbol{y}=\boldsymbol{x^i}+\alpha_0 \cdot (\boldsymbol{x^i}-\boldsymbol{x^j}) \oplus \boldsymbol{L\acute{e}vy(\beta)}$ \\
UNIF & $\boldsymbol{y}=\boldsymbol{x^i}+C \cdot (\boldsymbol{x^j}-\boldsymbol{x^k}) \oplus \boldsymbol{Unif(-1,1)}$ \\
NORM & $\boldsymbol{y}=\boldsymbol{x^i}+C \cdot (\boldsymbol{x^j}-\boldsymbol{x^k}) \oplus \boldsymbol{N(0,1)}$ \\
CONST & $\boldsymbol{y}=\boldsymbol{x^i}+F \cdot (\boldsymbol{x^j}-\boldsymbol{x^k})$ \\ \hline
\end{tabular}
\end{table}

\begin{table}
\centering
\caption{Reference points used on five datasets}
\label{tab:ref}
\begin{tabular}{cc}
\hline
Dataset & Reference Point (return, risk) \\ \hline
Hangseng & (0.0026, 0.0048) \\
DAX 100 & (0.0019, 0.0028) \\
FTSE 100 & (0.0024,0.0028) \\
S\&P 100 & (0.0018,0.0031) \\
Nikkei & (-0.0026,0.0017) \\ \hline
\end{tabular}
\end{table}

\begin{figure}
    \centerline{\includegraphics[width=0.7\textwidth]{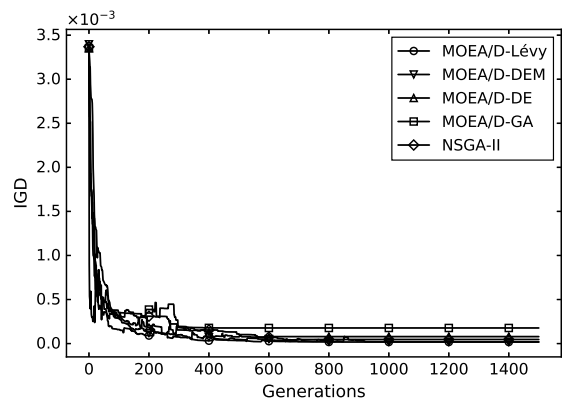}}
    \vspace*{8pt}
    \caption{IGD by generations on Nikkei in Experiment I} \label{fig:igdnikkeiI}
\end{figure}

\subsection{Experiment I: Comparison with Literature Methods} \label{sec:experiment-i-comparison-with-literature-methods}
In this experiment, MOEA/D-L\'evy and four comparison methods, MOEA/D-DEM, MOEA/D-DE, MOEA/D-GA and NSGA-II, are included. Among these five methods, four are based on MOEA/D framework described in Li’s study~\cite{HuiLi2009} (i.e., a small proportion to select parents from the whole population and an upper limitation for updating neighbors are set), while the decomposition method applied is NBI-style Tchebycheff approach~\cite{Zhang2010}. The selection method of all methods except the two GA-based algorithms is to select the current visited individual as one of the parents, and to randomly select the other individual as the other parents (i.e., same as Li’s study~\cite{HuiLi2009}). For MOEA/D-GA, all the parents are randomly selected from neighbors or the whole population, and a binary tournament selection is applied in NSGA-II. The mutation method in MOEA/D-L\'evy has been described in Line 12,~\algref{alg:moead-levy}. In MOEA/D-DEM, the mutation is the same as Li’s study~\cite{HuiLi2009} (i.e., DE mutation and polynomial mutation). In MOEA/D-DE, the mutation method is only based on DE mutation. This setting is applied in Zhang’s study~\cite{Zhang2010}. For two GA-based algorithms, SBX crossover and polynomial mutation are applied. This setting is the same as original NSGA-II~\cite{Deb2002} and MOEA/D~\cite{QingfuZhang2007} algorithms. In~\tabref{tab:exp1}, detailed mutation methods in Experiment I are listed.

Following the pre-experimental tuning described in \textbf{Section~\ref{sec:experimental-method}}, we obtained the following parameter values. In MOEA/D-L\'evy, $\alpha_0$ and $\beta$ of LF mutation are set to 1e-05 and 0.3. In MOEA/D-DEM and MOEA/D-DE, scaling factor $F$ of DE mutation is set to 1.3. In MOEA/D-L\'evy and MOEA/D-DEM, mutation rate is set to 1/$n$ (i.e., $n$ is the size of dataset). In MOEA/D-GA and NSGA-II, crossover rate is set to 0.7. In MOEA/D-GA, mutation rate is 0.05 and in NSGA-II, mutation rate is 0.01.

\subsection{Experiment II: Comparison with Other Distribution-based Mutation Methods} \label{sec:experiment-ii-comparison-with-other-distribution-based-mutation-methods}
In this experiment, mutation methods based on four probability distributions, namely L\'evy-stable distribution (LEVY), uniform distribution (UNIF), standard normal distribution (NORM) and constant (CONST) are compared to show the effectiveness of LF. Among these four methods, MOEA/D framework and selection method in Li’s study~\cite{HuiLi2009} are applied. The mutation operators of these methods are similar to DE mutation, but the scaling factors are drawn from the above mentioned probability distributions. In addition, two parents are selected in LEVY, while three are selected in the other three mutation methods. As the goal of this experiment is to show how LF contributes to optimization, no polynomial mutation is applied in all methods. \tabref{tab:exp2} presents the detailed formula of the mutation methods in Experiment II.

Following the pre-experimental tuning method described in \textbf{Section~\ref{sec:experimental-method}}, we obtained the following parameter values. Parameter $C$ is set to 1.0 and 0.5 and in UNIF and NORM, respectively. The parameters of LF mutation in LEVY are the same as that of MOEA/D-L\'evy in \textbf{Section~\ref{sec:experiment-i-comparison-with-literature-methods}}. The parameters of CONST is the same as that of MOEA/D-DE in \textbf{Section~\ref{sec:experiment-i-comparison-with-literature-methods}}. It is interesting to notice that CONST and MOEA/D-DE are the same method.

\begin{figure}
    \centerline{\includegraphics[width=0.7\textwidth]{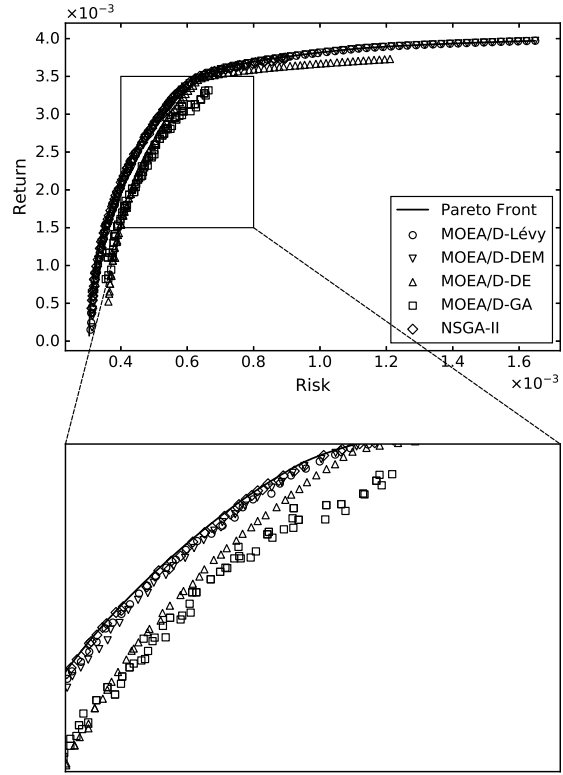}}
    \vspace*{8pt}
    \caption{Final population and zoom-in on Nikkei in Experiment I}
    \label{fig:popnikkeiIcom}
\end{figure}

\begin{figure*}
    \centering
    \begin{minipage}{0.5\textwidth}
        \includegraphics[width=\textwidth]{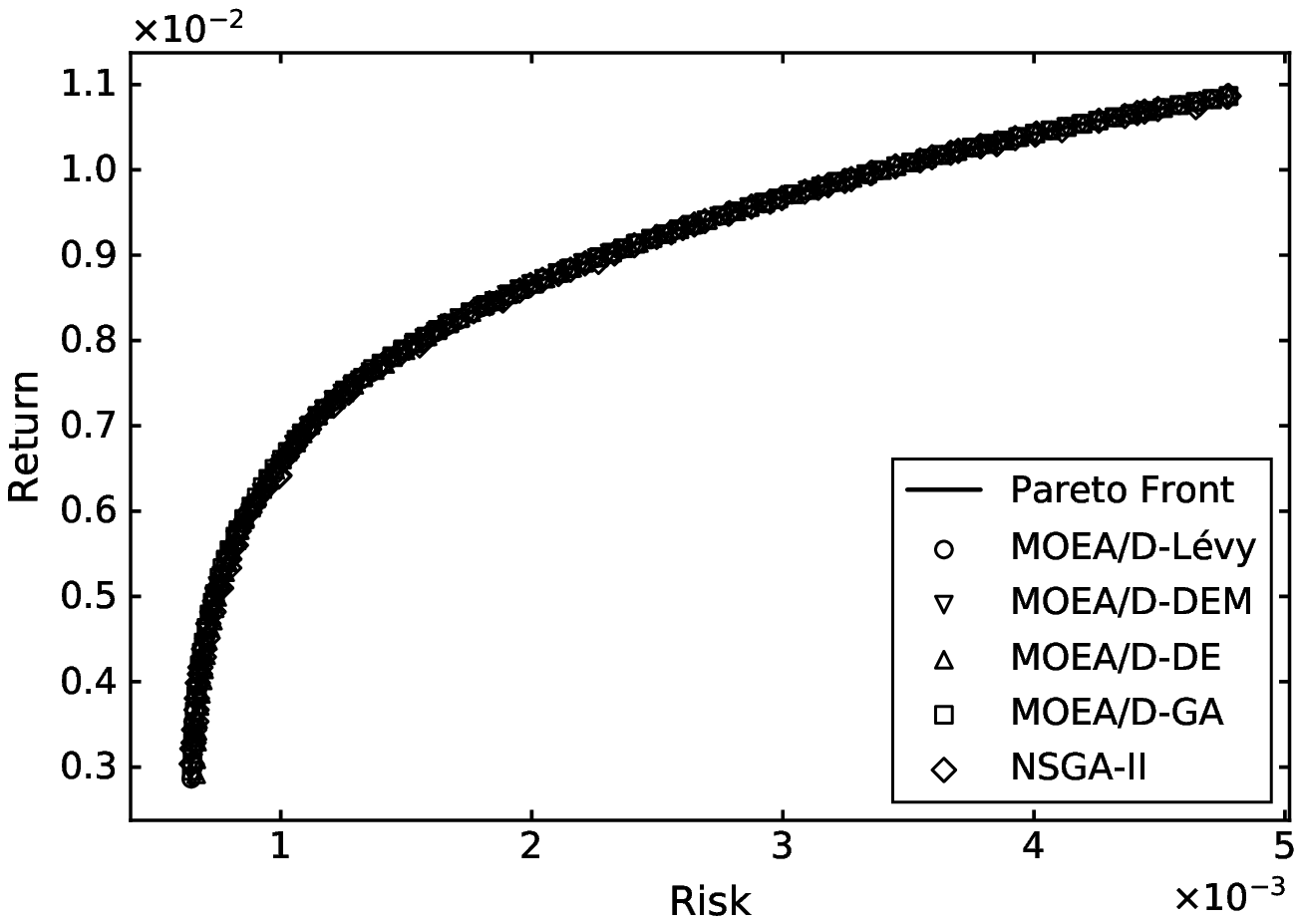}
        \subcaption{Hangseng}
        \label{fig:finpopexp1a}
    \end{minipage}%
    \begin{minipage}{0.5\textwidth}
        \includegraphics[width=\textwidth]{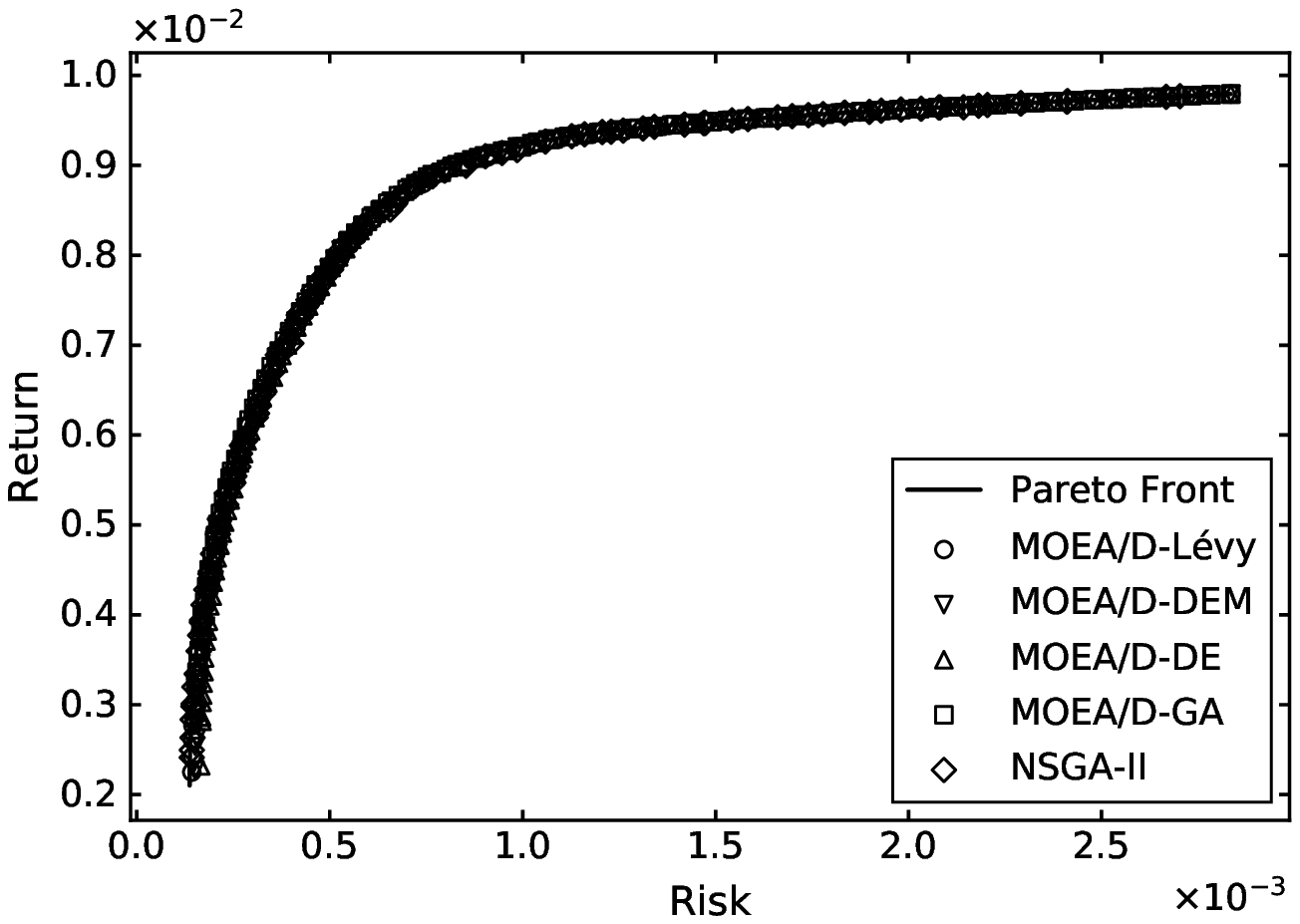}
        \subcaption{DAX 100}
        \label{fig:finpopexp1b}
    \end{minipage}%
    \quad
    \begin{minipage}{0.5\textwidth}
        \includegraphics[width=\textwidth]{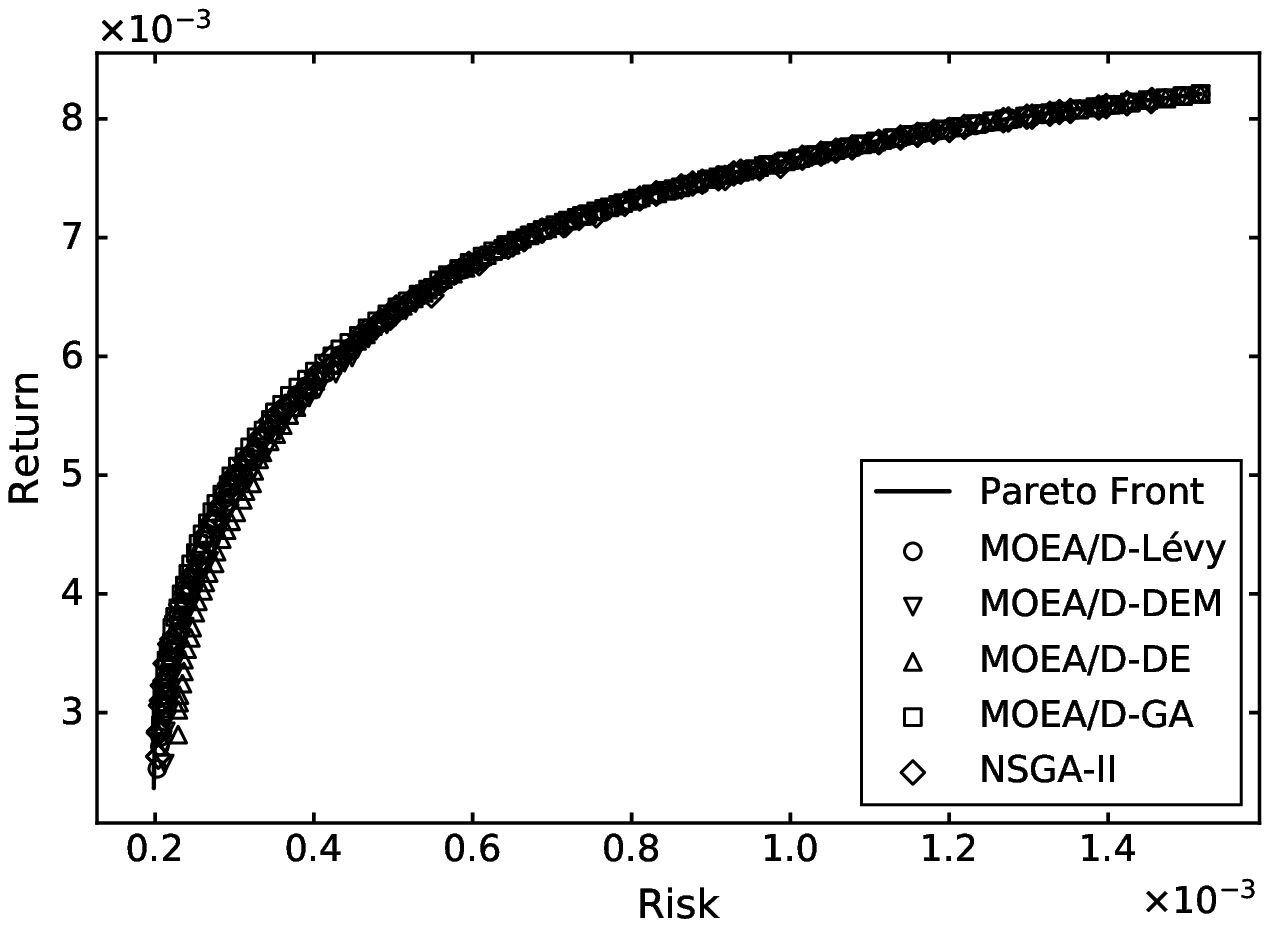}
        \subcaption{FTSE 100}
        \label{fig:finpopexp1c}
    \end{minipage}%
    \begin{minipage}{0.5\textwidth}
        \includegraphics[width=\textwidth]{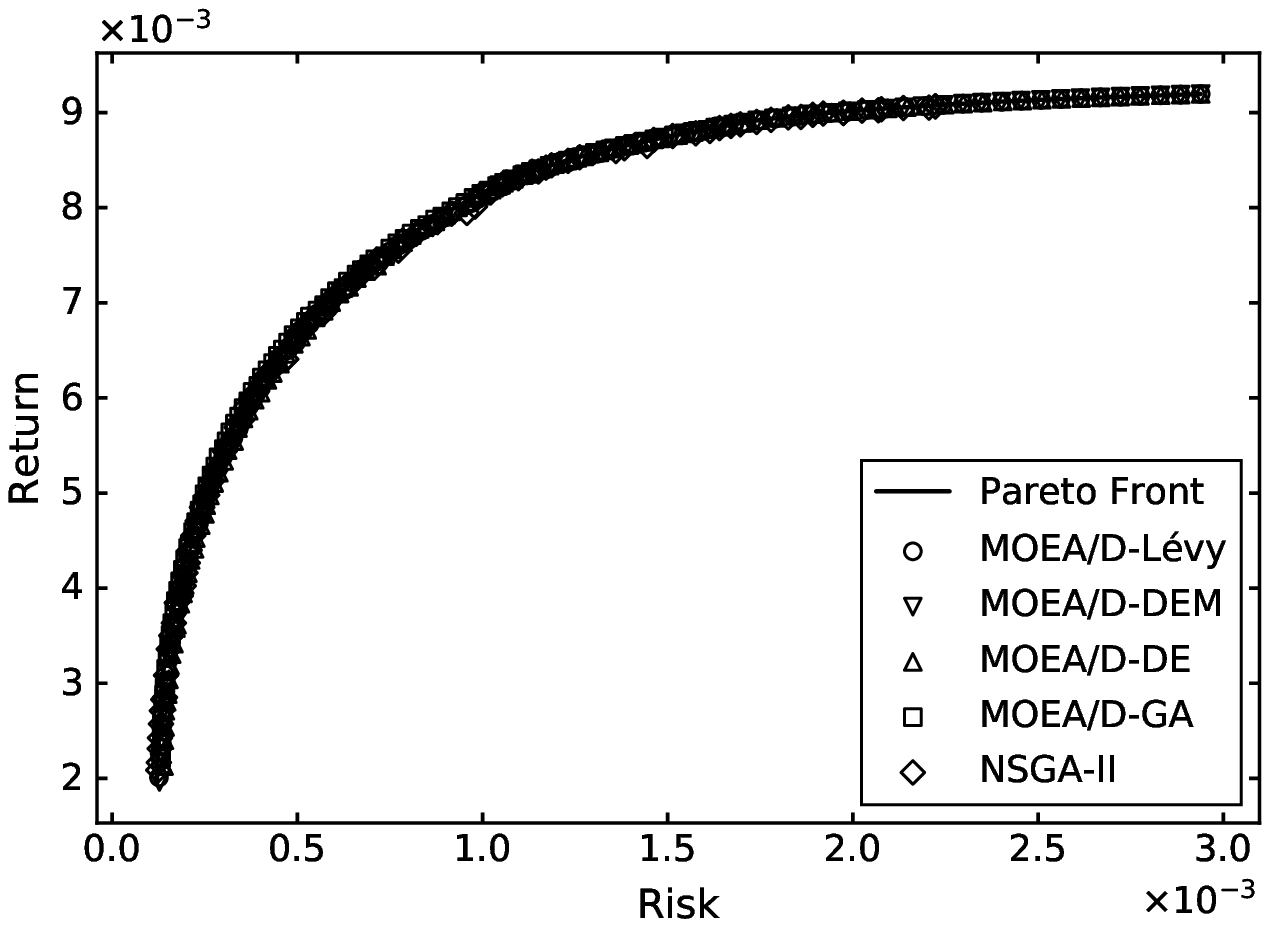}
        \subcaption{S\&P 100}
        \label{fig:finpopexp1d}
    \end{minipage}%
    \quad
    \begin{minipage}{0.5\textwidth}
        \includegraphics[width=\textwidth]{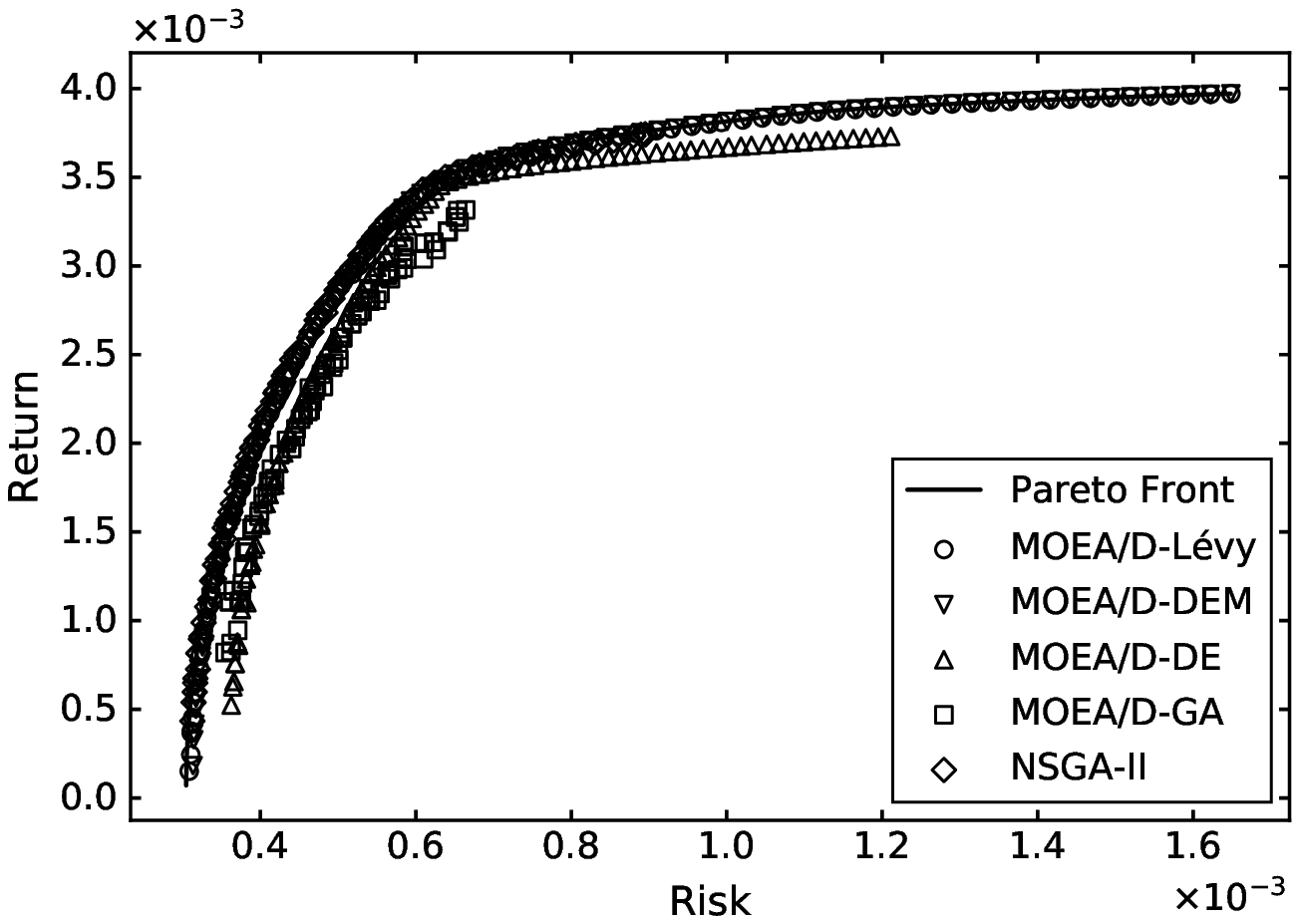}
        \subcaption{Nikkei}
        \label{fig:finpopexp1e}
    \end{minipage}%
    \vspace*{8pt}
    \caption{Final population on five datasets in objective space (Experiment I)}
    \label{fig:finpopexp1}
\end{figure*}

\begin{figure*}
    \centering
    \begin{minipage}{0.5\textwidth}
        \includegraphics[width=\textwidth]{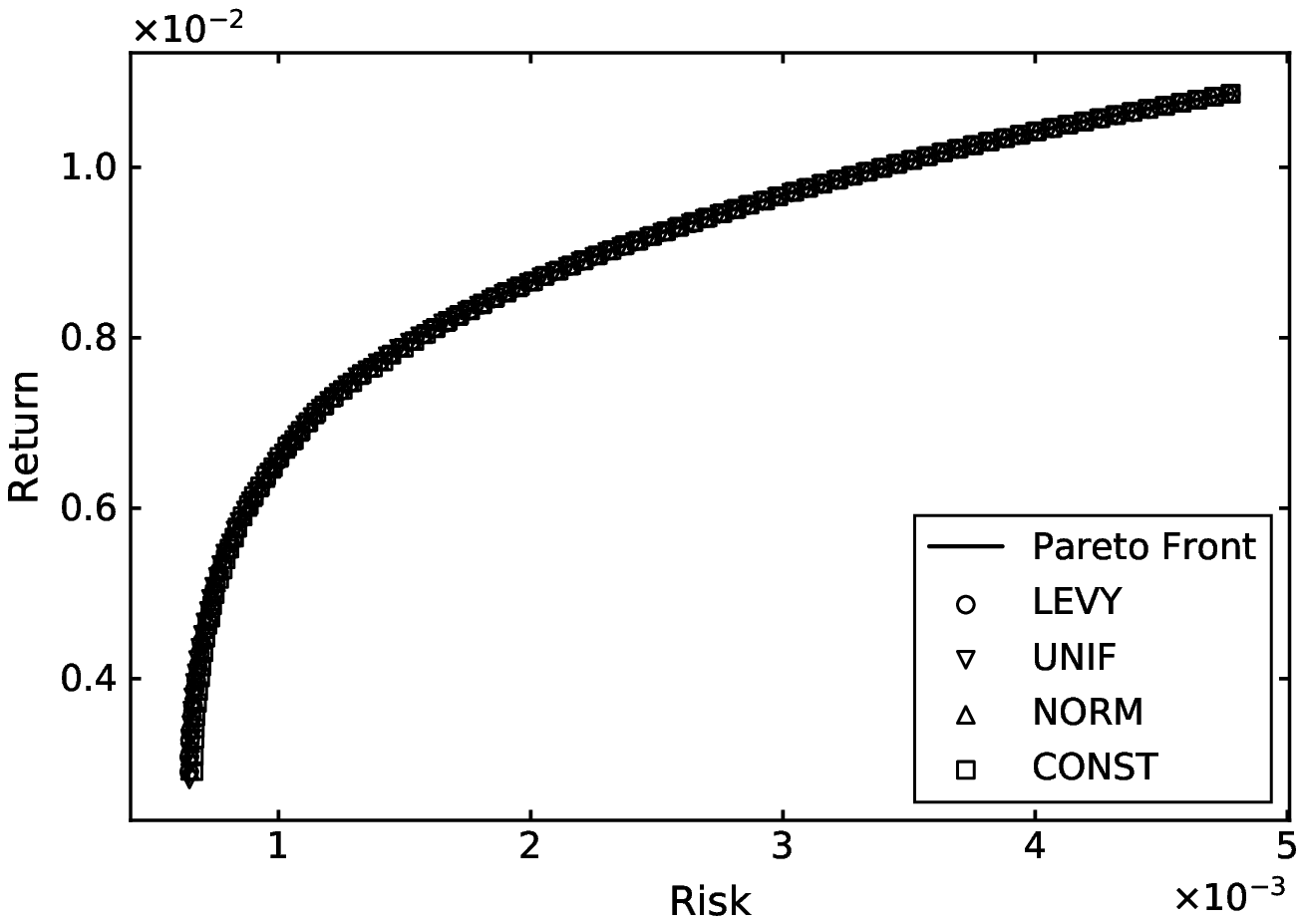}
        \subcaption{Hangseng}
        \label{fig:finpopexp2a}
    \end{minipage}%
    \begin{minipage}{0.5\textwidth}
        \includegraphics[width=\textwidth]{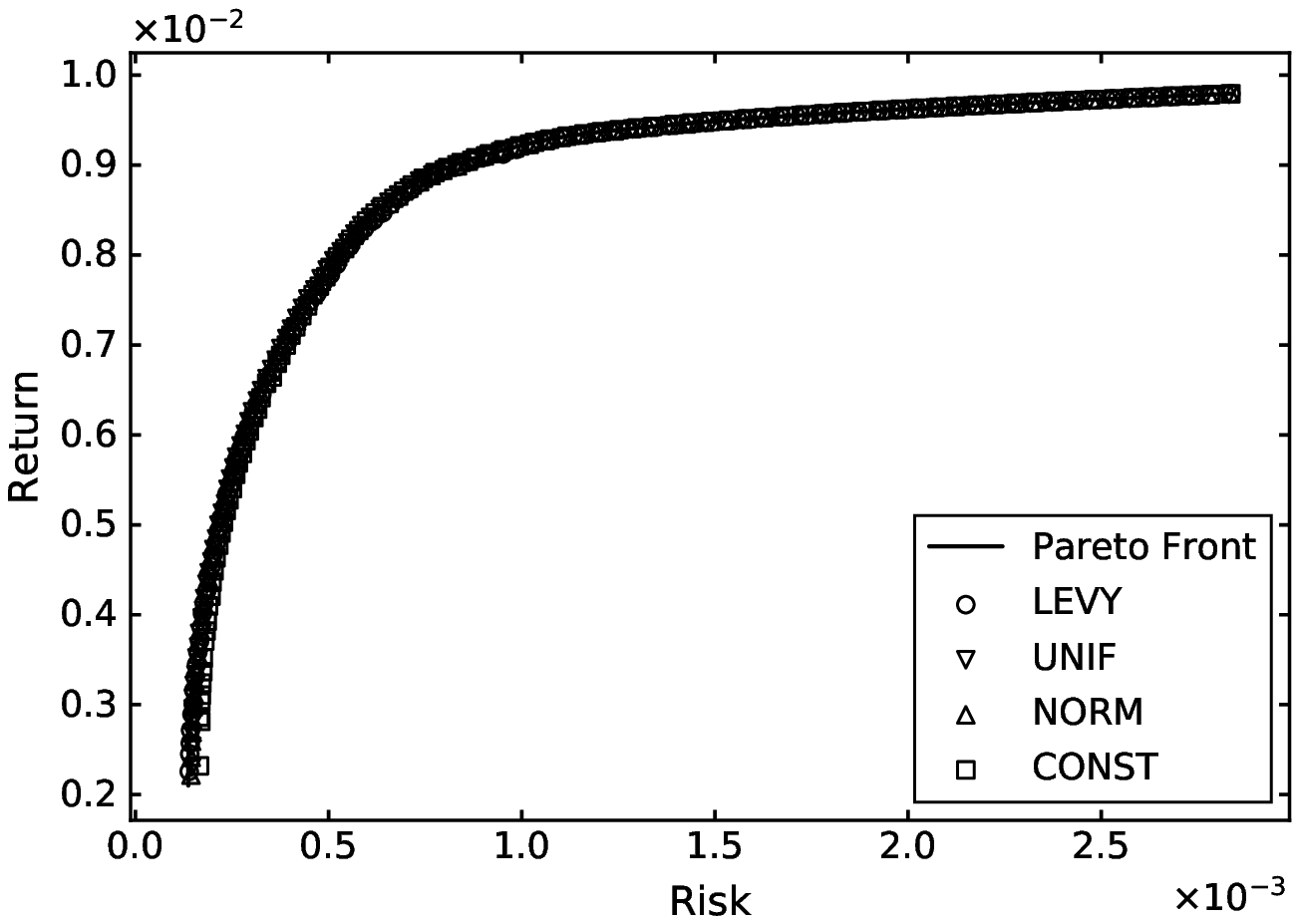}
        \subcaption{DAX 100}
        \label{fig:finpopexp2b}
    \end{minipage}%
    \quad
    \begin{minipage}{0.5\textwidth}
        \includegraphics[width=\textwidth]{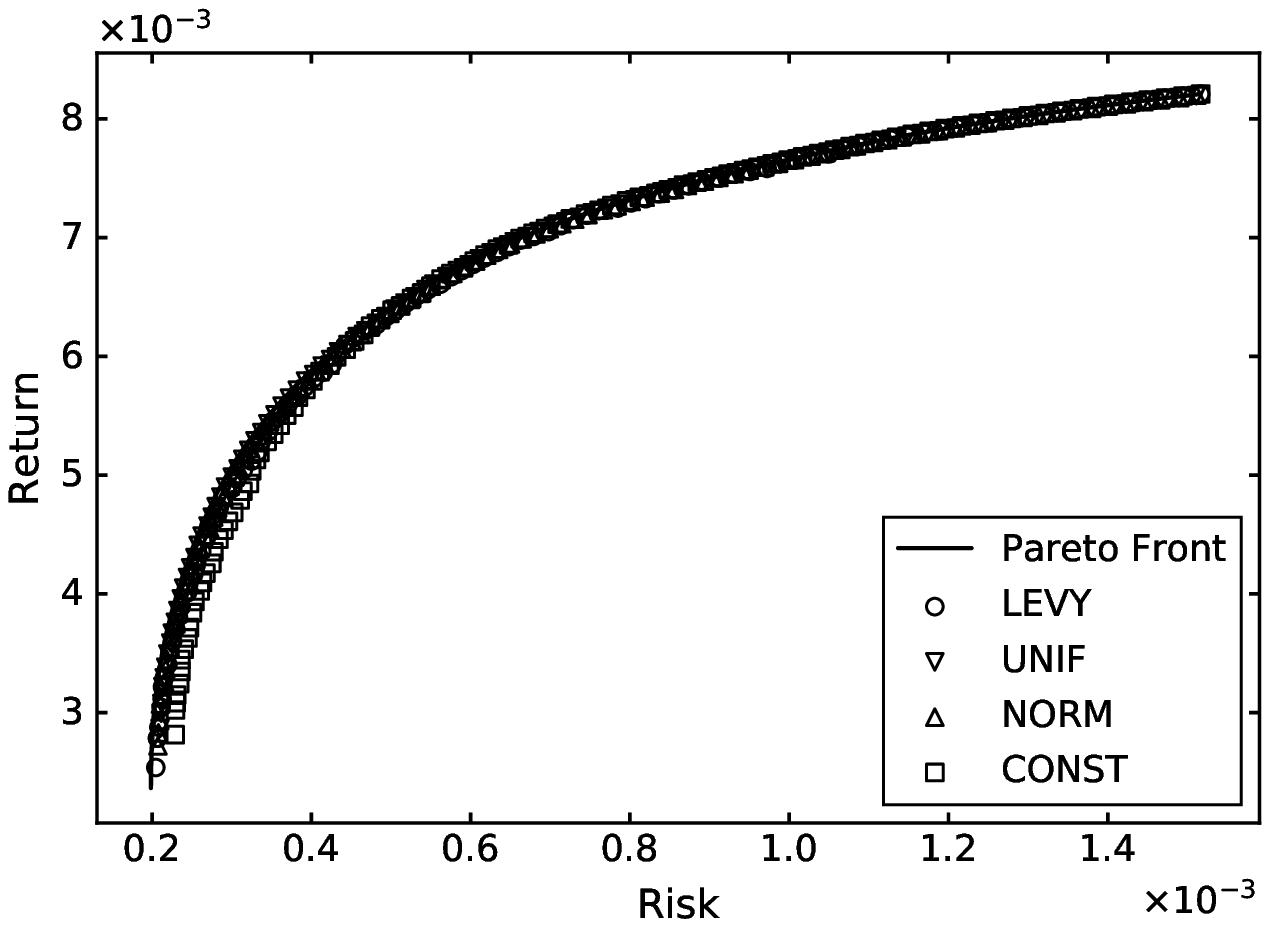}
        \subcaption{FTSE 100}
        \label{fig:finpopexp2c}
    \end{minipage}%
    \begin{minipage}{0.5\textwidth}
        \includegraphics[width=\textwidth]{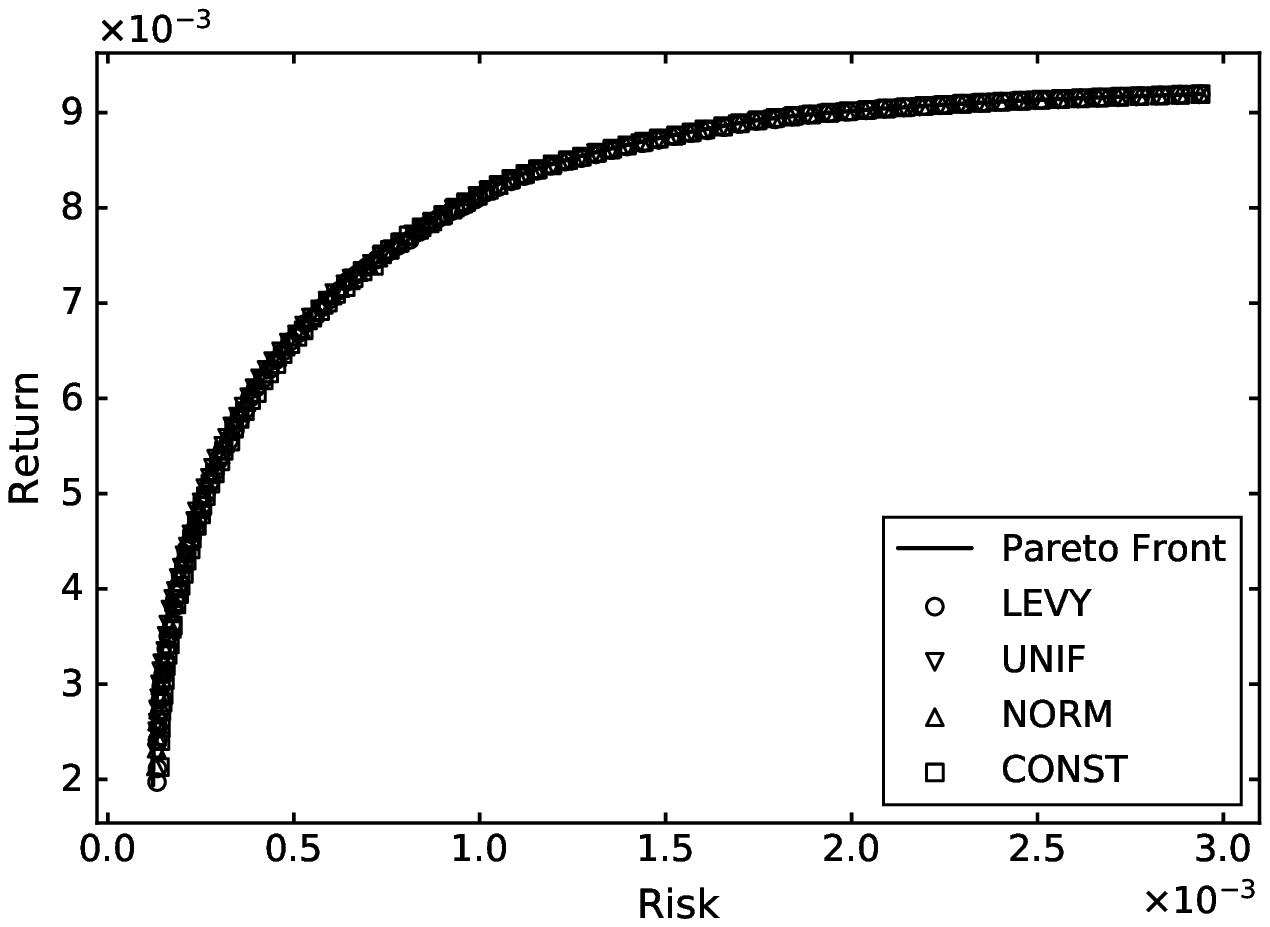}
        \subcaption{S\&P 100}
        \label{fig:finpopexp2d}
    \end{minipage}%
    \quad
    \begin{minipage}{0.5\textwidth}
        \includegraphics[width=\textwidth]{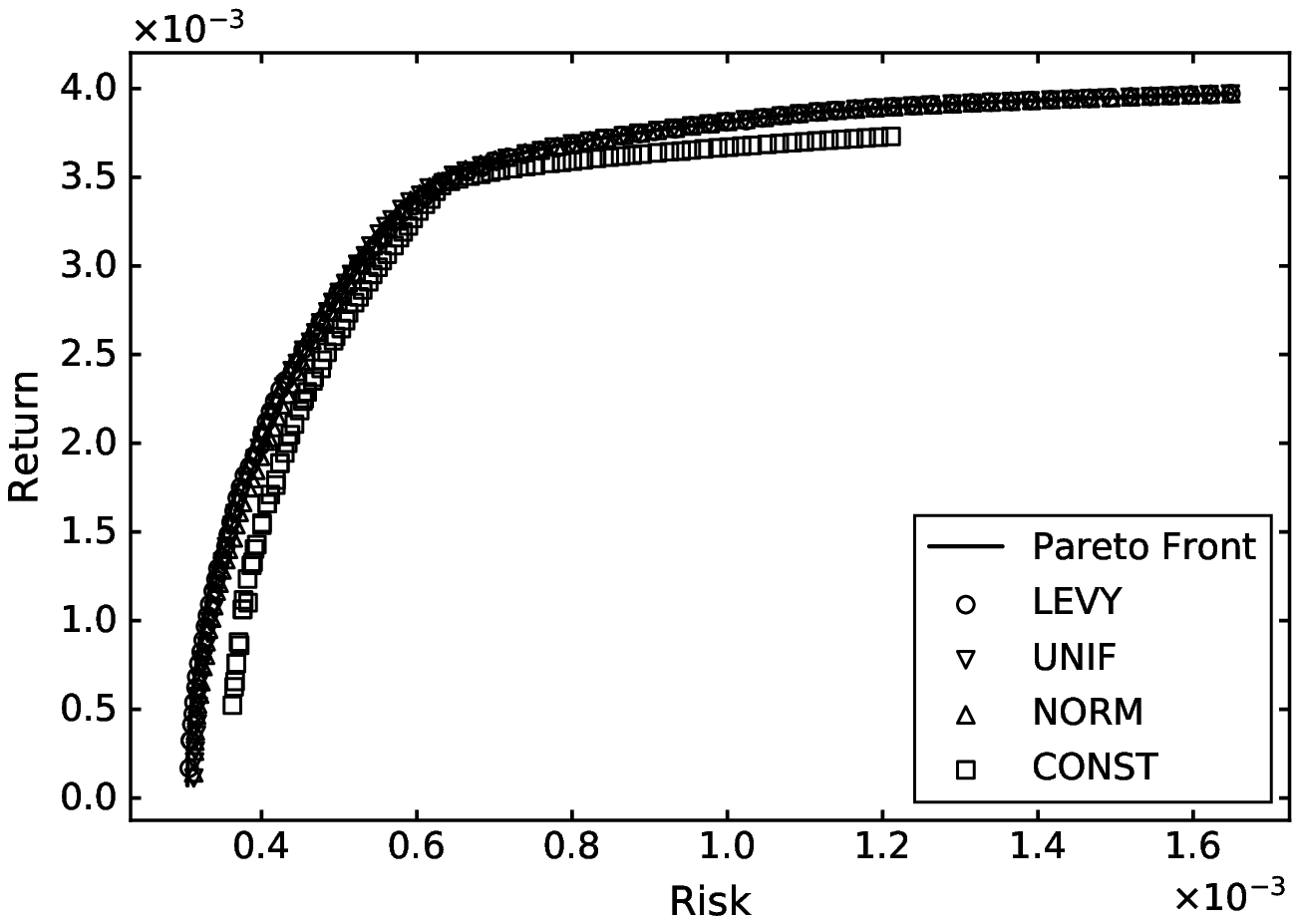}
        \subcaption{Nikkei}
        \label{fig:finpopexp2e}
    \end{minipage}%
    \vspace*{8pt}
    \caption{Final population on five datasets in objective space (Experiment II)}
    \label{fig:finpopexp2}
\end{figure*}

\begin{figure*}
    \centering
    \begin{minipage}{0.5\textwidth}
        \includegraphics[width=\textwidth]{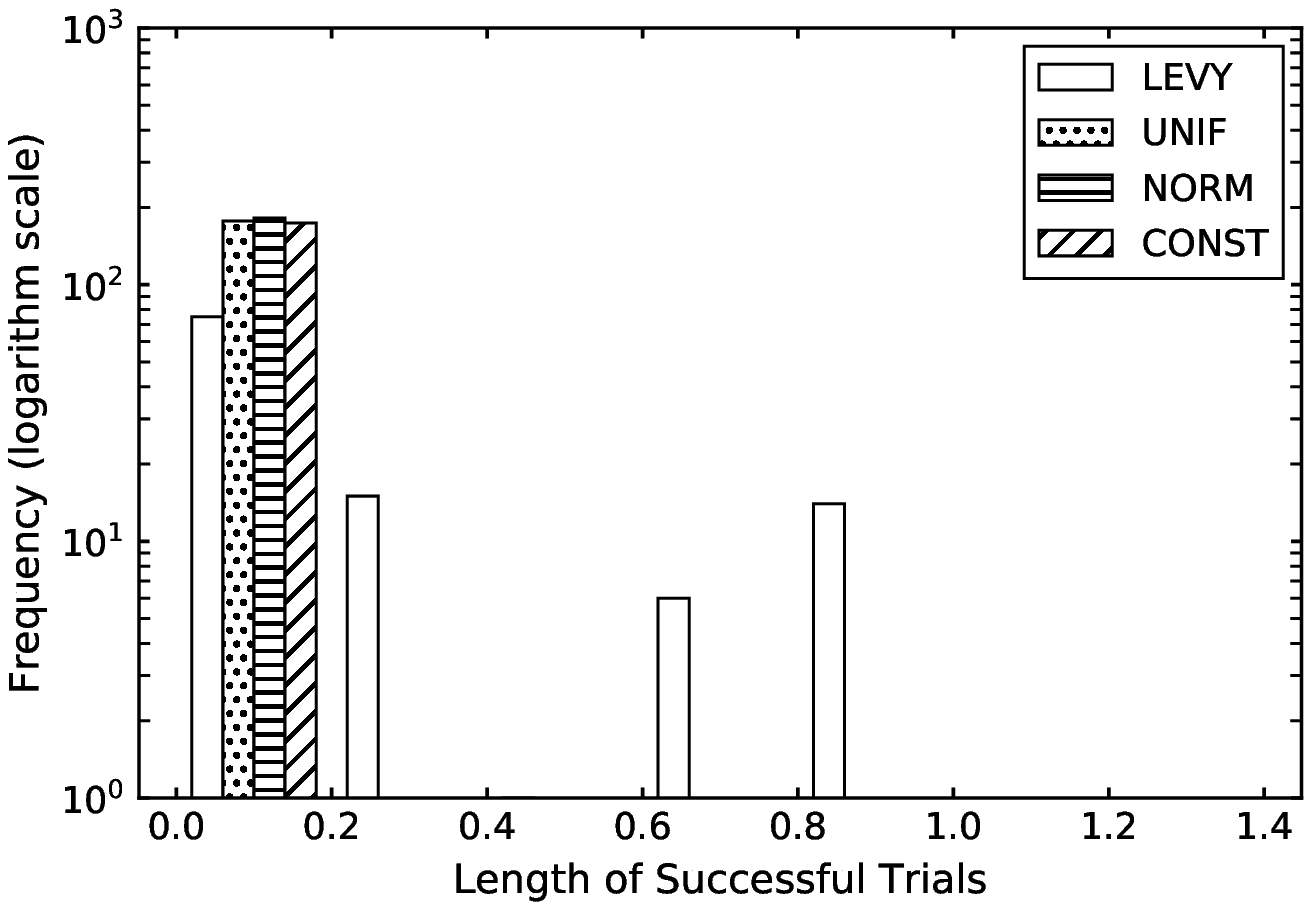}
        \subcaption{Successful trials}
        \label{fig:trialpopexp2gen1a}
    \end{minipage}%
    \begin{minipage}{0.5\textwidth}
        \includegraphics[width=\textwidth]{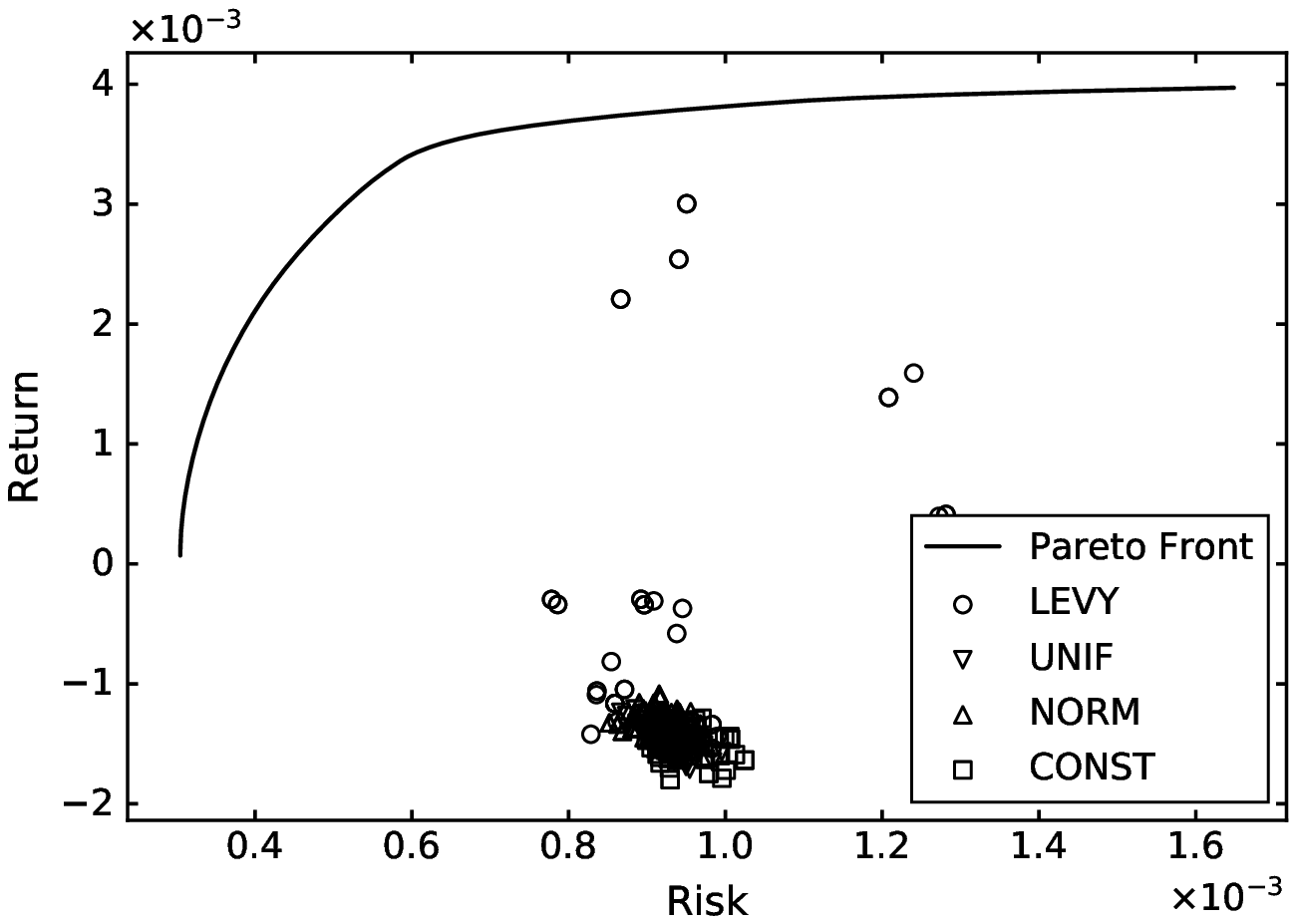}
        \subcaption{Population in objective space}
        \label{fig:trialpopexp2gen1b}
    \end{minipage}
    \vspace*{8pt}
    \caption{Experiment II, Nikkei Dataset (1st generation) Left: Frequency of ``Succesful trials" (when the mutation operator generates an offspring that is better than its parent) against the length of the mutation step. Right: population in the objective space.}
    \label{fig:trialpopexp2gen1}
\end{figure*}
\begin{figure*}
    \centering
    \begin{minipage}{0.5\textwidth}
        \includegraphics[width=\textwidth]{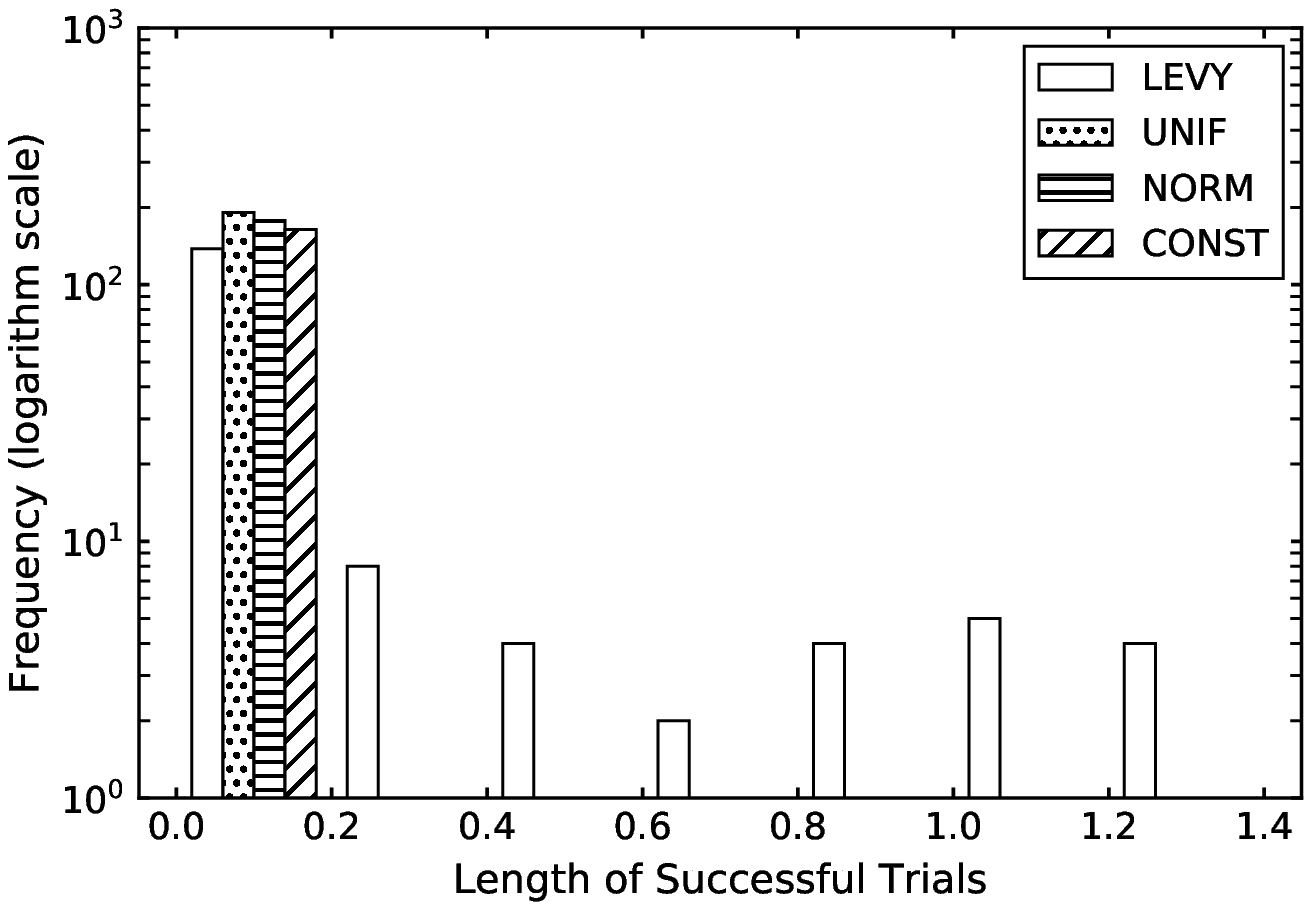}
        \subcaption{Successful trials}
        \label{fig:trialpopexp2gen3a}
    \end{minipage}%
    \begin{minipage}{0.5\textwidth}
        \includegraphics[width=\textwidth]{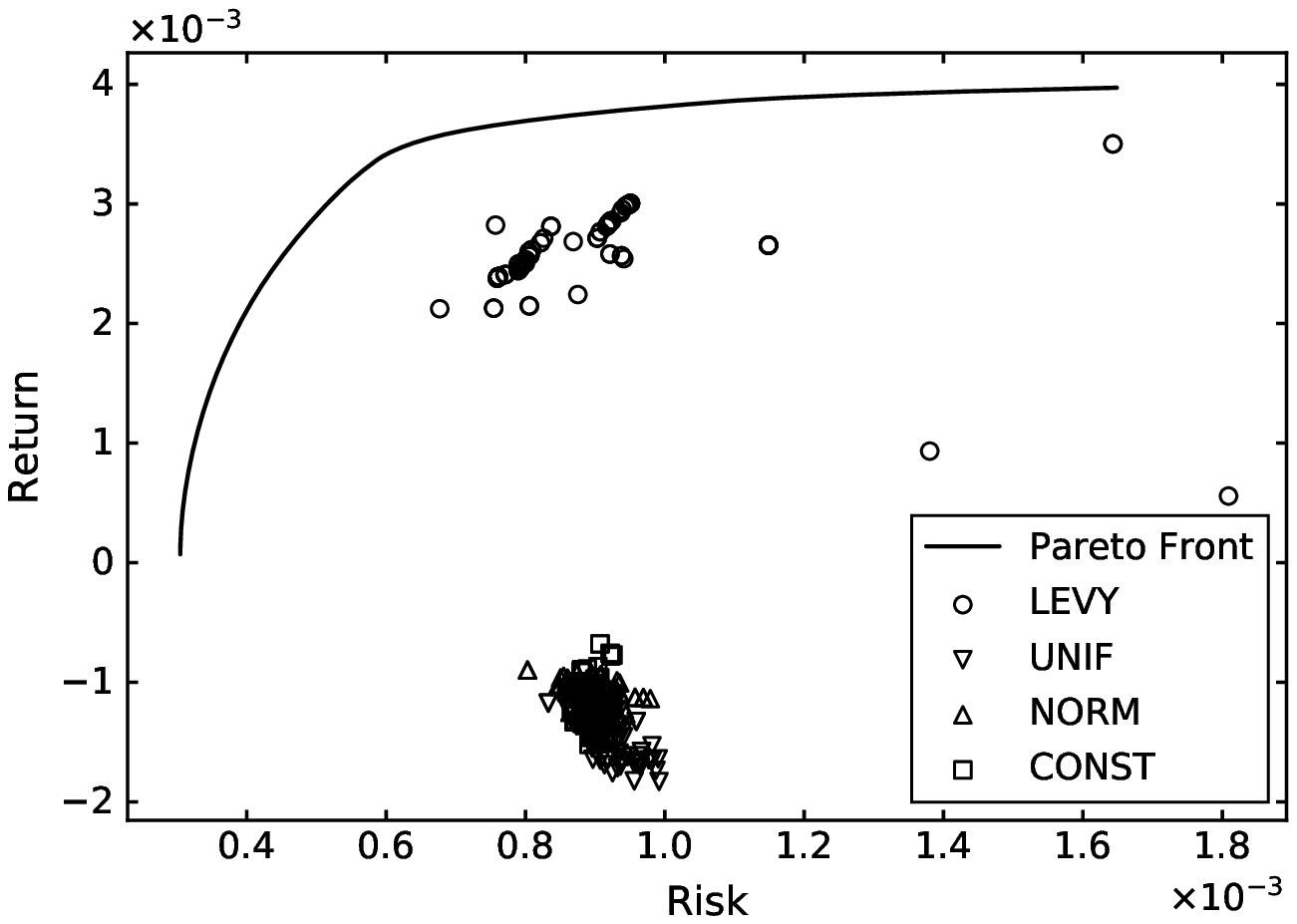}
        \subcaption{Population in objective space}
        \label{fig:trialpopexp2gen3b}
    \end{minipage}
    \vspace*{8pt}
    \caption{Experiment II, Nikkei Dataset (3rd generation) Left: Frequency of ``Succesful trials" (when the mutation operator generates an offspring that is better than its parent) against the length of the mutation step. Right: population in the objective space.}
    \label{fig:trialpopexp2gen3}
\end{figure*}
\begin{figure*}
    \centering
    \begin{minipage}{0.5\textwidth}
        \includegraphics[width=\textwidth]{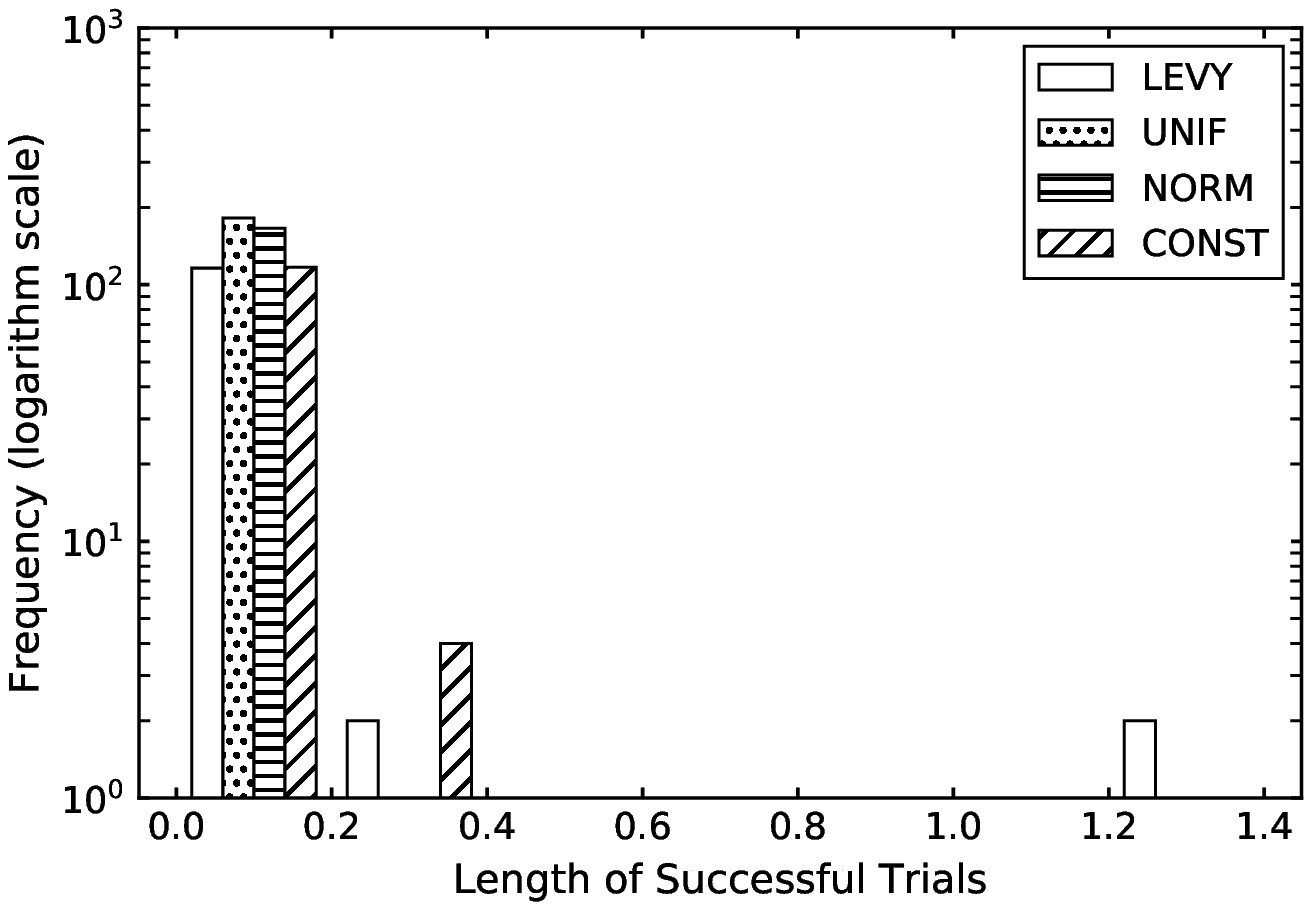}
        \subcaption{Successful trials}
        \label{fig:trialpopexp2gen10a}
    \end{minipage}%
    \begin{minipage}{0.5\textwidth}
        \includegraphics[width=\textwidth]{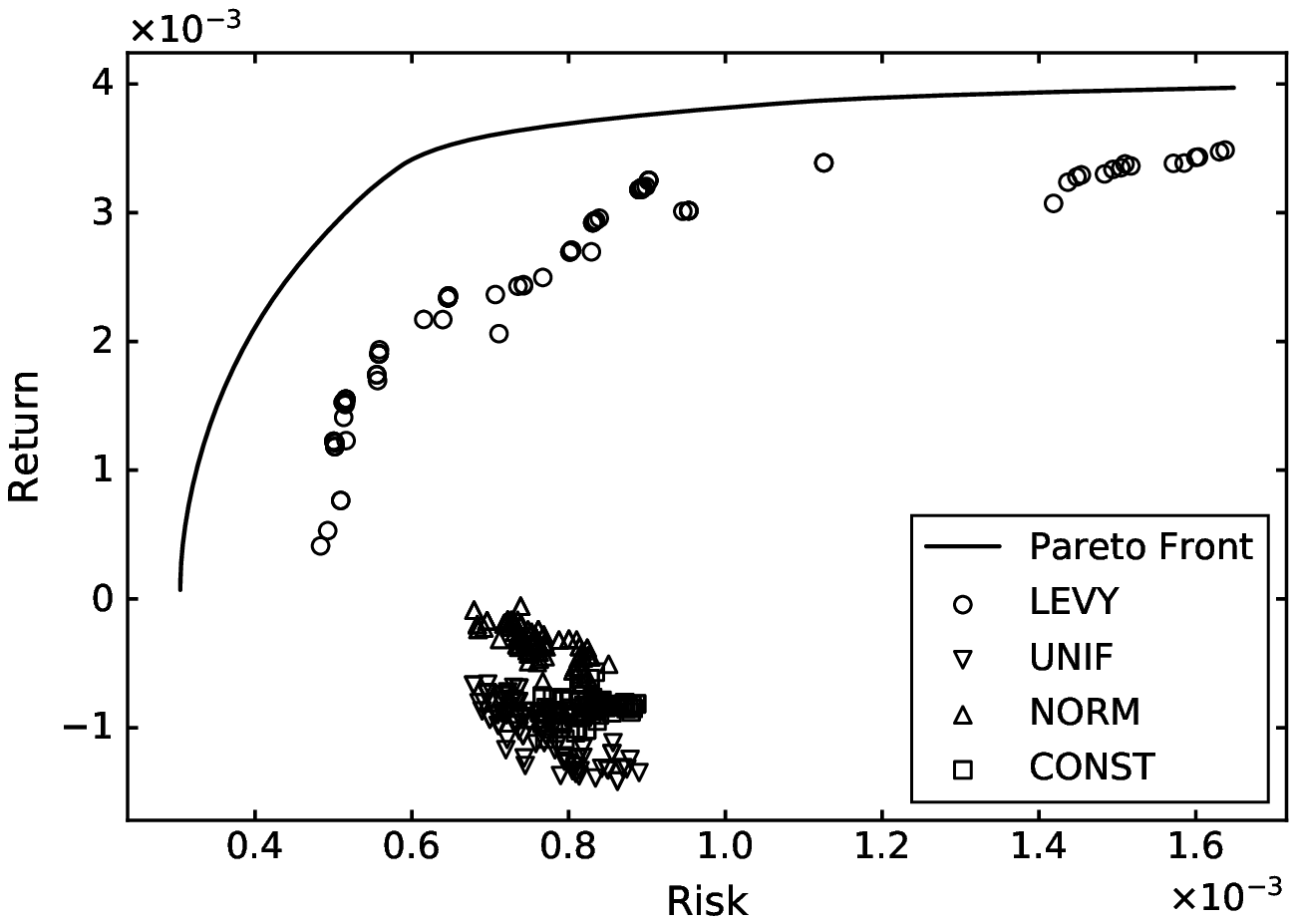}
        \subcaption{Population in objective space}
        \label{fig:trialpopexp2gen10b}
    \end{minipage}
    \vspace*{8pt}
    \caption{Experiment II, Nikkei Dataset (10th generation) Left: Frequency of ``Succesful trials" (when the mutation operator generates an offspring that is better than its parent) against the length of the mutation step. Right: population in the objective space..}
    \label{fig:trialpopexp2gen10}
\end{figure*}

\begin{figure*}
    \centering
    \begin{minipage}{0.5\textwidth}
        \includegraphics[width=\textwidth]{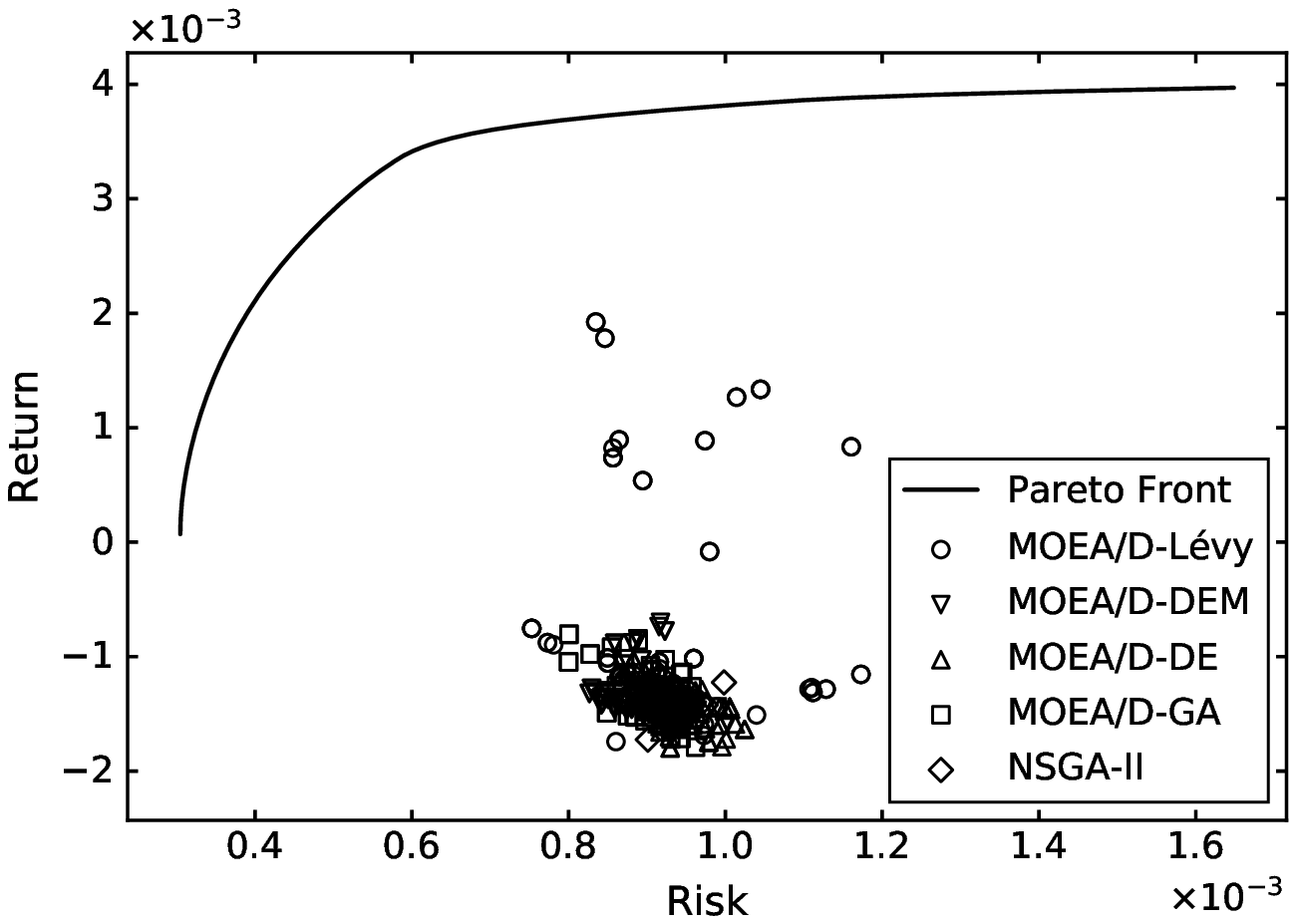}
        \subcaption{1st generation}
        \label{fig:popexp1a}
    \end{minipage}%
    \begin{minipage}{0.5\textwidth}
        \includegraphics[width=\textwidth]{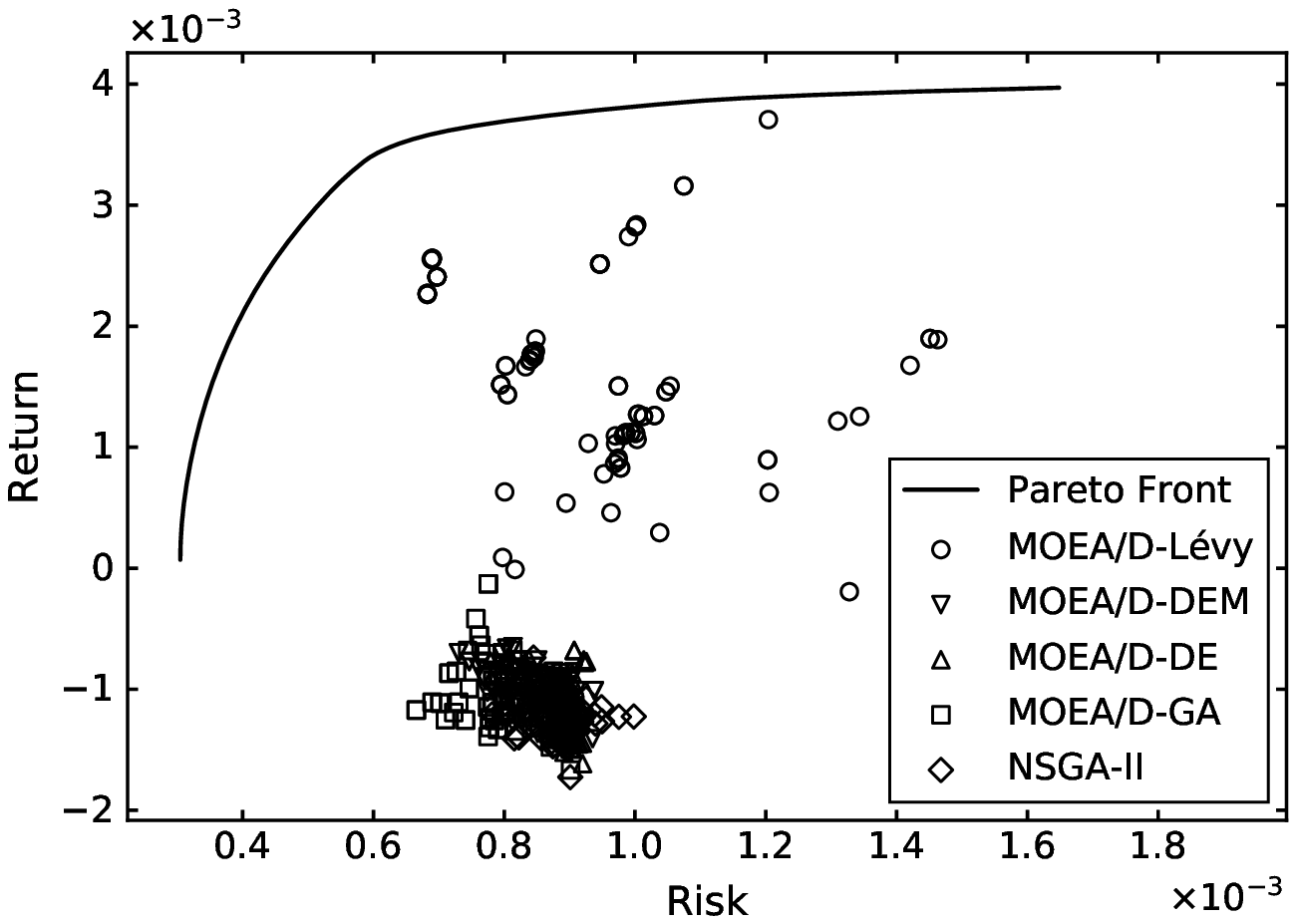}
        \subcaption{3rd generation}
        \label{fig:popexp1b}
    \end{minipage}%
    \quad
    \begin{minipage}{0.5\textwidth}
        \includegraphics[width=\textwidth]{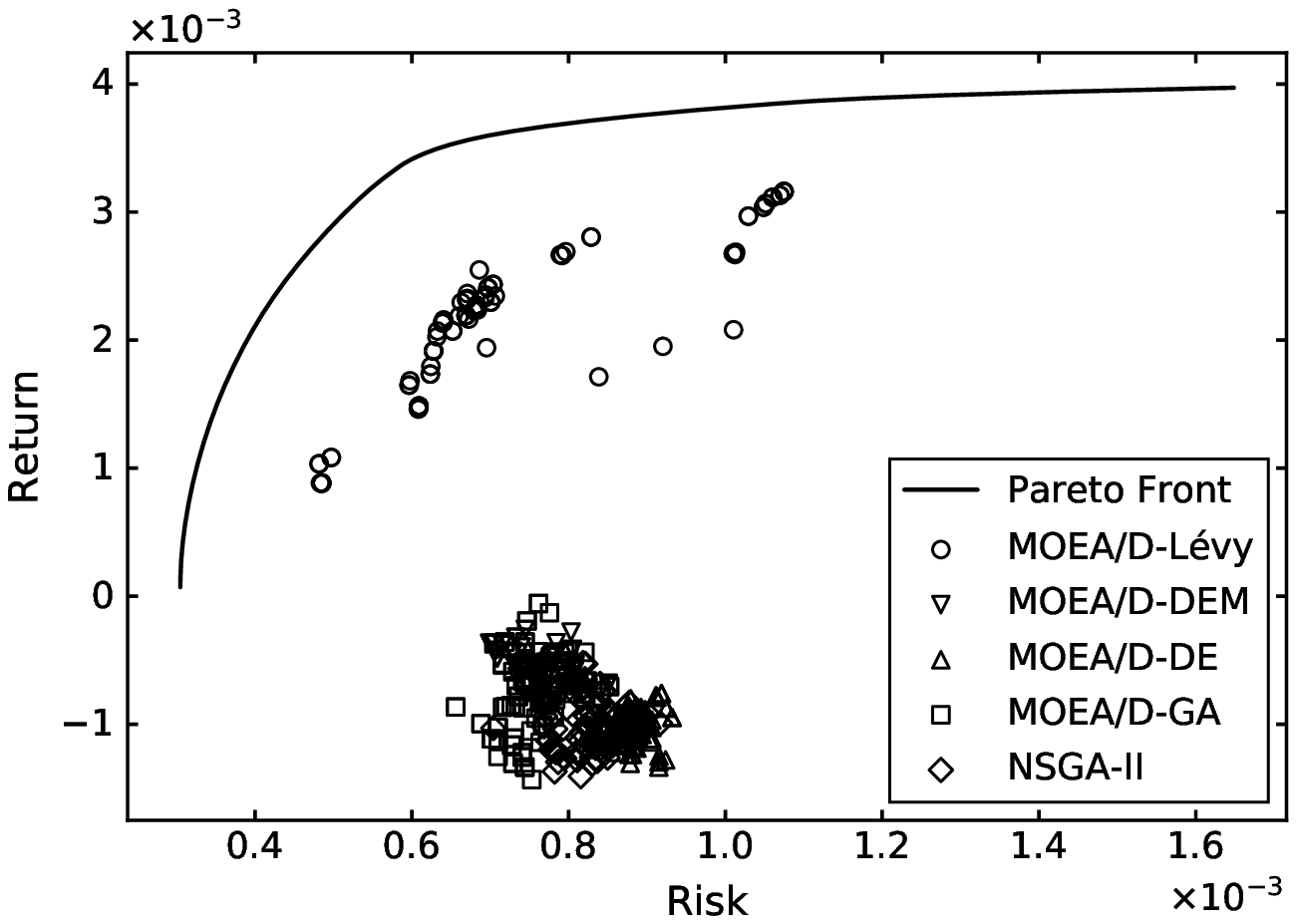}
        \subcaption{5th generation}
        \label{fig:popexp1c}
    \end{minipage}%
    \begin{minipage}{0.5\textwidth}
        \includegraphics[width=\textwidth]{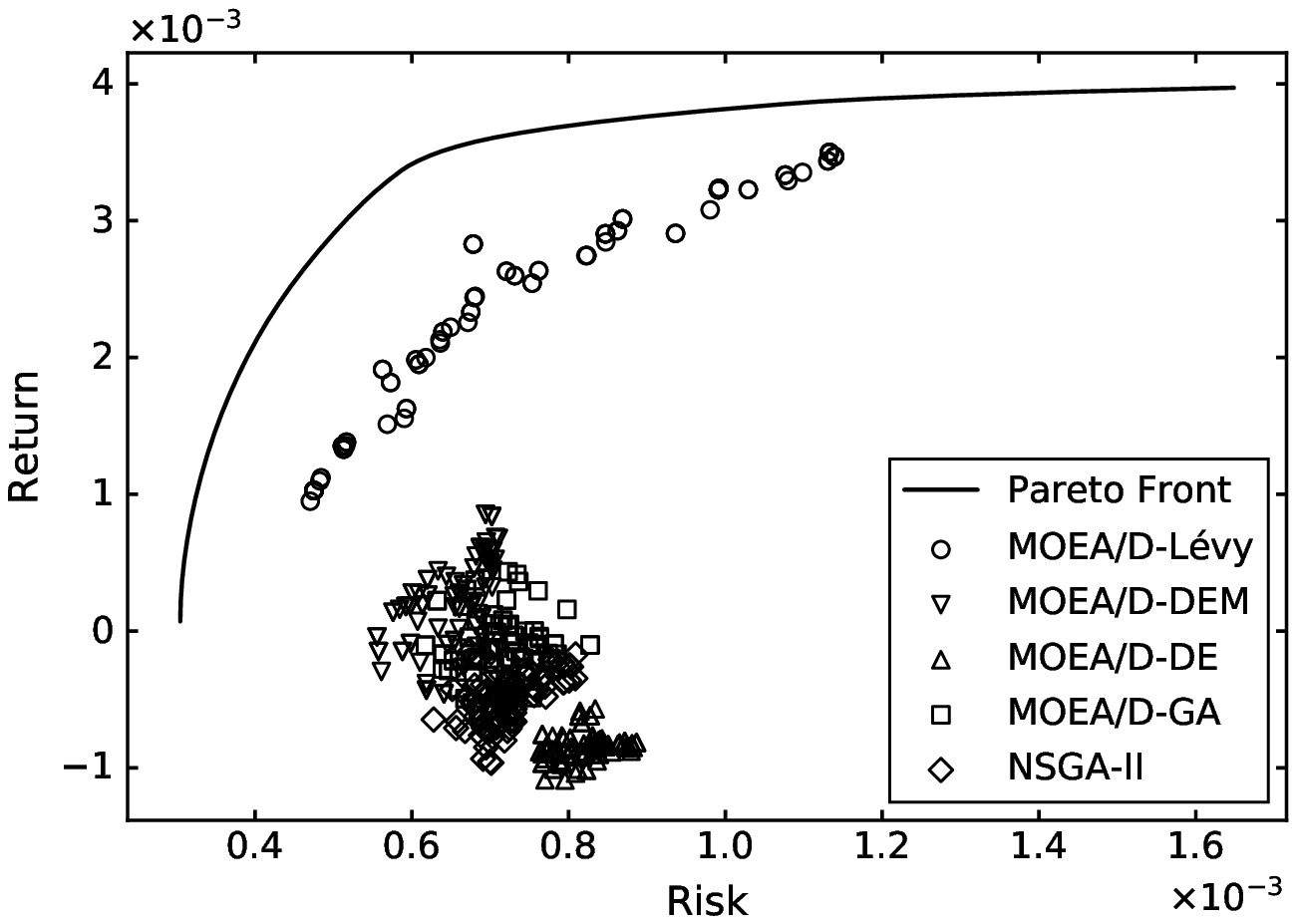}
        \subcaption{10th generation}
        \label{fig:popexp1d}
    \end{minipage}%
    \quad
    \begin{minipage}{0.5\textwidth}
        \includegraphics[width=\textwidth]{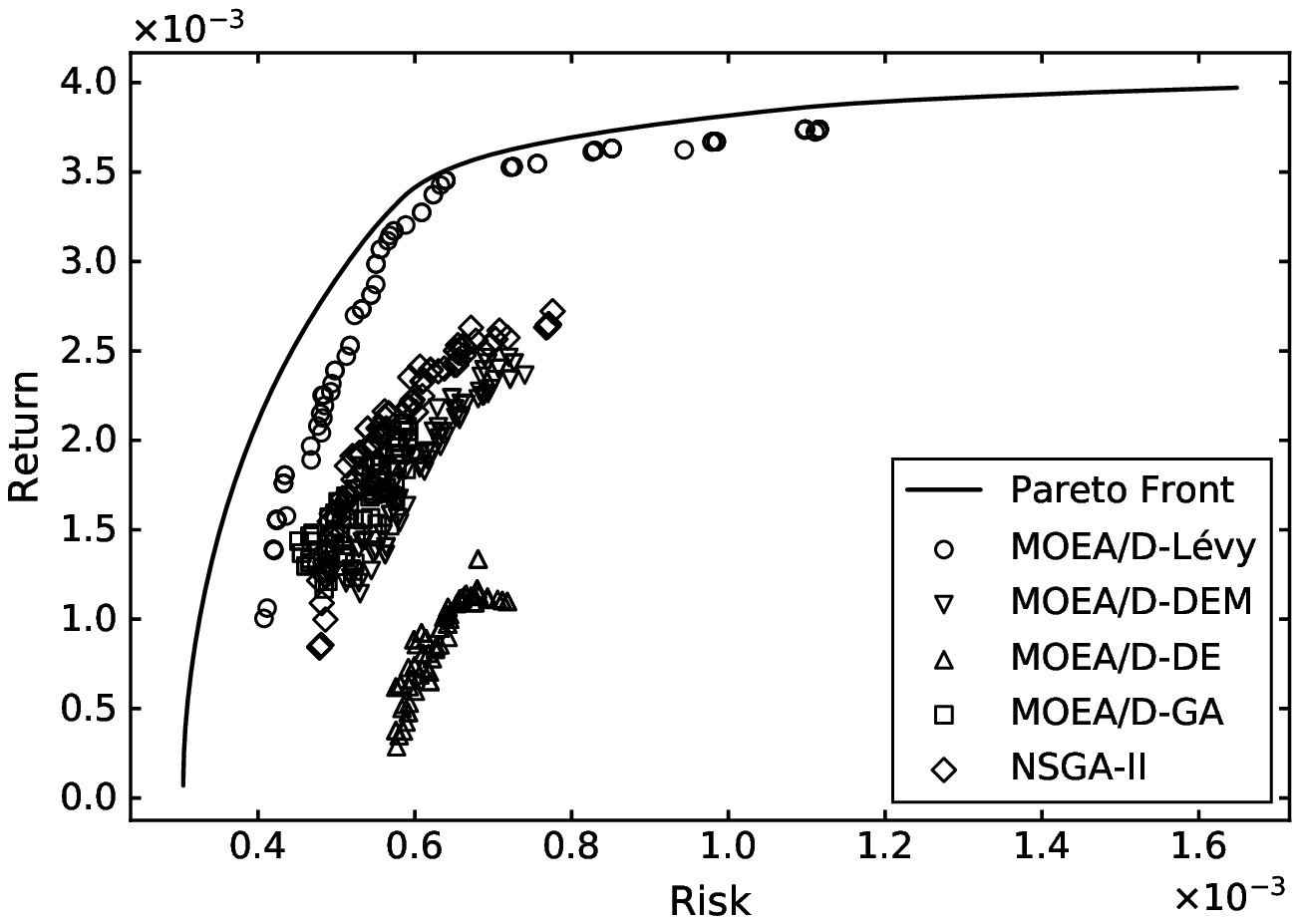}
        \subcaption{50th generation}
        \label{fig:popexp1e}
    \end{minipage}%
    \begin{minipage}{0.5\textwidth}
        \includegraphics[width=\textwidth]{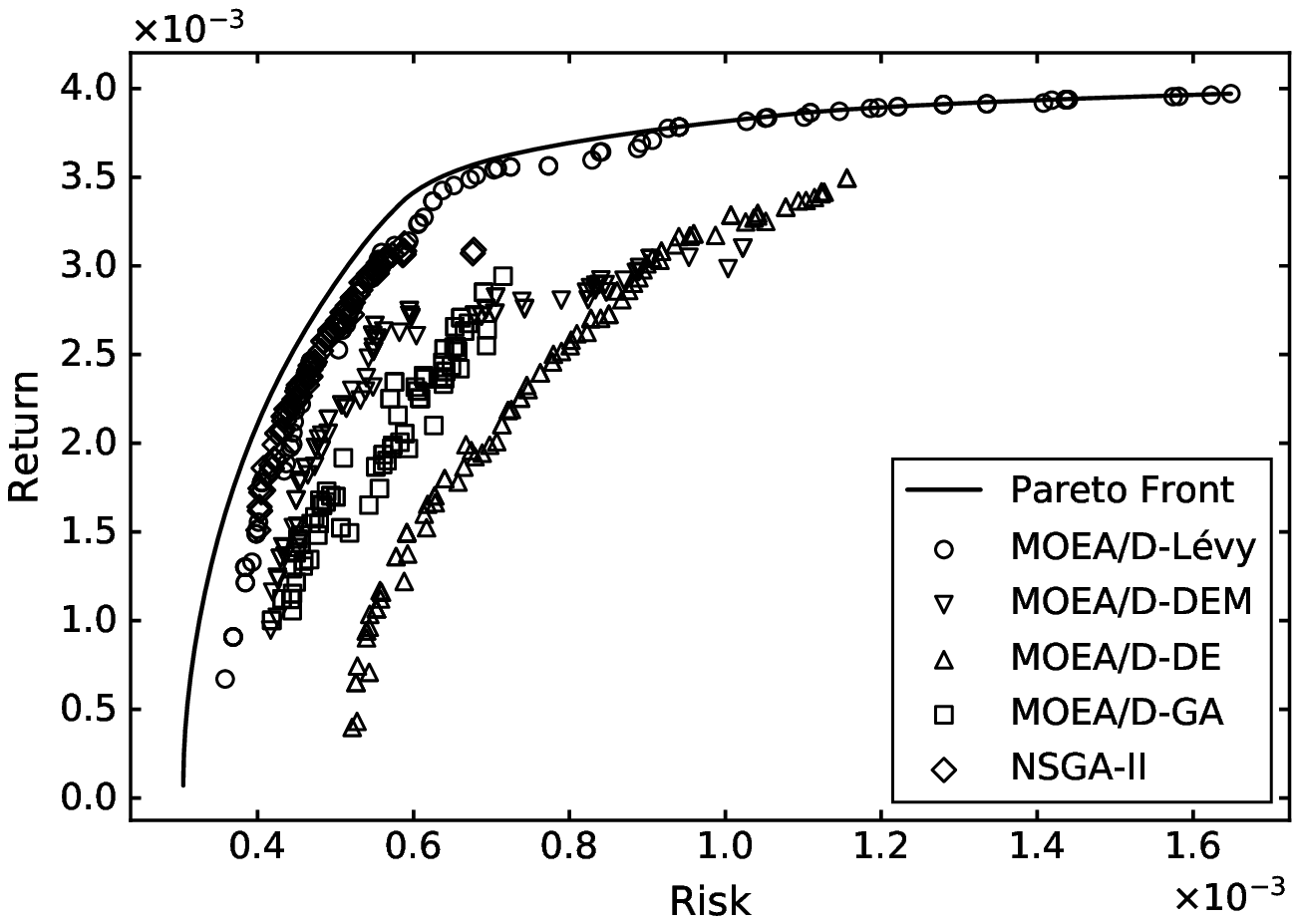}
        \subcaption{100th generation}
        \label{fig:popexp1f}
    \end{minipage}
    \vspace*{8pt}
    \caption{Experiment I: Populations of different algorithms over generations in the Nikkei dataset. MOEA/D-L\'evy covers a wider area early in the optimization, which leads to a better distribution over the Pareto Front around the 100th generation.}
    \label{fig:popexp1}
\end{figure*}

\begin{figure}
    \centerline{\includegraphics[width=0.7\textwidth]{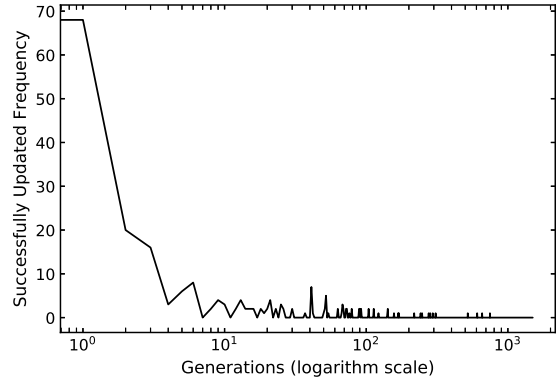}}
    \vspace*{8pt}
    \caption{Successfully updated long trials by generations}
    \label{fig:trajbygen}
\end{figure}

\begin{figure}
    \centerline{\includegraphics[width=0.5\textwidth]{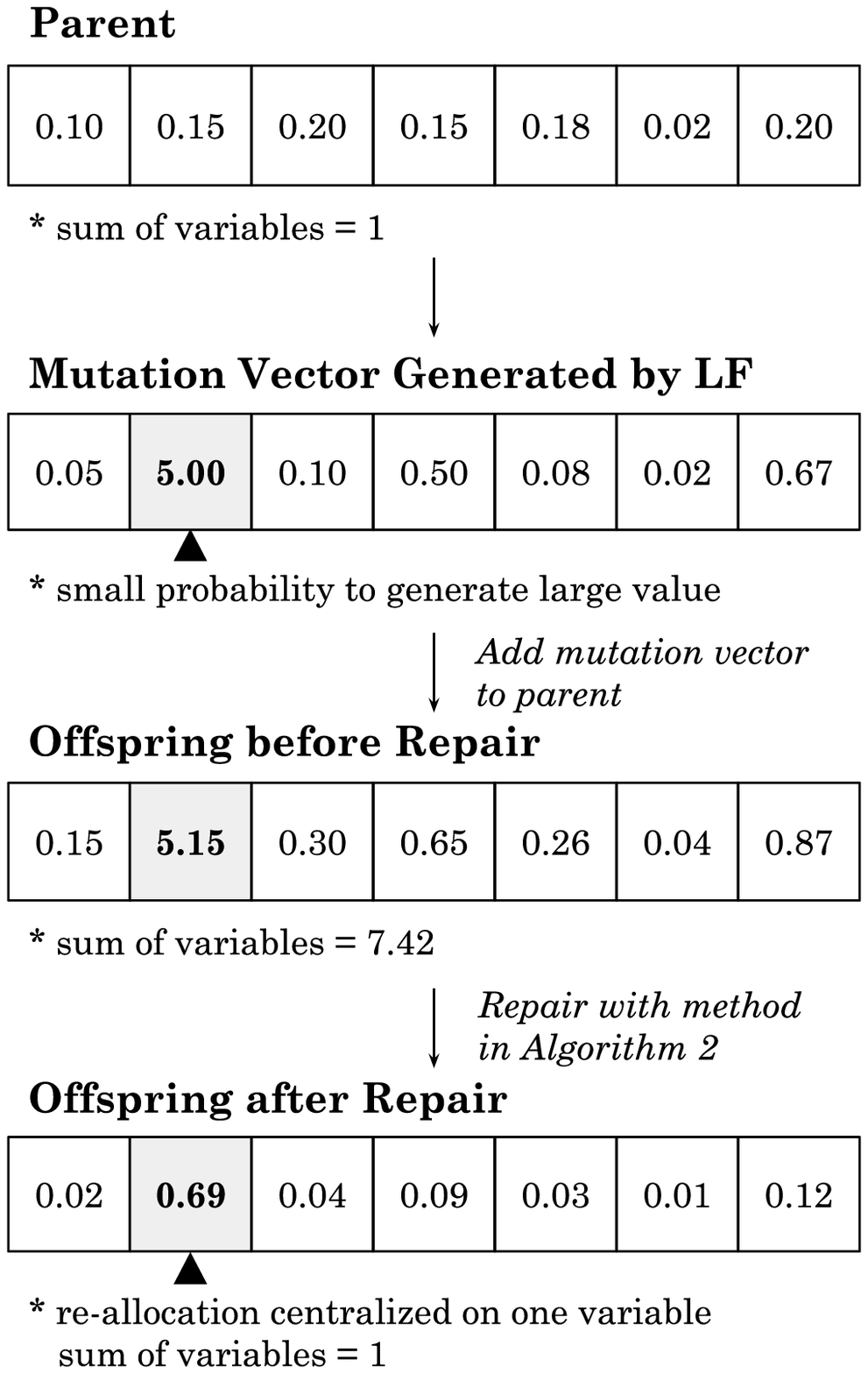}}
    \vspace*{8pt}
    \caption{How LF mutation helps exploration in PO. An interaction between
    long trajectories and unit constraint leads to candidates with high allocation in only some assets.}
    \label{fig:poexplan}
\end{figure}

\subsection{Experimental Results} \label{sec:experimental-results}
The reference points used to compute HV are reported in~\tabref{tab:ref}. The numerical results of both experiments on the five datasets are presented from~\tabref{tab:res1hangseng} to~\tabref{tab:res2nikkei}. The medians of the best algorithm in each metric are in bold font. In addition, those best medians decorated with underlines indicate that the corresponding algorithms perform better than algorithms with second best medians, through a Wilcoxon Rank Sum Test with a significant level of 5\%. As an example,~\figref{fig:igdnikkeiI} shows IGD changing by generations on Nikkei in Experiment I (best run of IGD). \figref{fig:popnikkeiIcom} illustrates the final population on the same dataset and experiment, as well as a zoom-in. The final population of other datasets and experiments are shown in~\figref{fig:finpopexp1} and~\ref{fig:finpopexp2}. The results for each dataset are detailed as follows.

\begin{itemize}
    \item \textbf{Hangseng}: In Experiment I, MOEA/D-L\'evy holds the best median in terms of MS, $\Delta$ and IGD, while MOEA/D-GA performs best on GD, S, and HV. Both methods show a statistical significance compared with the methods holding the second-best median. In Experiment II, NORM performs best on GD, MS, IGD, and HV. CONST holds the best S, and UNIF holds the best $\Delta$.
    \item \textbf{DAX 100}: In Experiment I, MOEA/D-L\'evy shows a statistically significant superiority in terms of MS, $\Delta$, IGD and HV, while MOEA/D-GA performs better in GD and S with a statistical significance. In Experiment II, NORM is the best method in terms of five metrics except for S, especially on GD, MS and IGD, the statistical test shows a significant difference. For the S metric, CONST performs best.
    \item \textbf{FTSE 100}: In Experiment I, MOEA/D-L\'evy shows a clear superiority on MS, $\Delta$, IGD and HV, while MOEA/D-GA and MOEA/D-DE perform best on GD and S, respectively. In Experiment II, LEVY performs best on MS, $\Delta$, IGD and HV. On MS and IGD, it shows a statistical significance in the comparison with NORM which holds second place. NORM performs best on GD, while CONST performs best on S.
    \item \textbf{S\&P 100}: In Experiment I, MOEA/D-L\'evy holds the best median on $\Delta$, IGD and HV. Especially, on $\Delta$ and HV, there is a significant difference between the second-best method. MOEA/D-GA, MOEA/D-DE, and MOEA/D-DEM are the best method in terms of GD, S, and MS, respectively. In Experiment II, UNIF holds the best GD and S, while NORM performs best on the rest metrics. LEVY holds the second place on MS, IGD, and HV, and shows a comparable performance.
    \item \textbf{Nikkei}: In Experiment I, MOEA/D-L\'evy performs best on $\Delta$, IGD and HV. Especially, on $\Delta$ and HV, there is a significant difference between the second-best method. NSGA-II, MOEA/D-DE, and MOEA/D-DEM are the best method in terms of GD, S, and MS, respectively. In Experiment II, LEVY shows a significant superiority in terms of GD, MS, $\Delta$, IGD and HV. For S, CONST performs best.
\end{itemize}

 In addition, it is hard to find a difference when considering the plots of the solution distributions. On FTSE 100, the MOEA/D-DE or CONST perform relatively worse on retrieving low-risk portfolio (i.e., they have fewer solutions in the bottom left regions in~\figref{fig:finpopexp1c} and~\ref{fig:finpopexp2c}). On S\&P 100 and Nikkei, the two GA-based algorithms only retrieve a part of the front (i.e., show in~\figref{fig:finpopexp1d} and~\ref{fig:finpopexp1e}). On Nikkei, the solution set retrieved by MOEA/D-DE or CONST is relatively narrow and far from the Pareto Front (i.e., show in~\figref{fig:finpopexp1e} and~\ref{fig:finpopexp2e}).

\begin{table*}[p]
\small\centering
\caption{Numerical results on Hangseng in Experiment I}
\label{tab:res1hangseng}
\begin{tabular}{ccccccc}
\hline
\multicolumn{2}{c}{\multirow{2}{*}{Metric}} & \multicolumn{4}{c}{MOEA/D} & \multirow{2}{*}{NSGA-II} \\ \cline{3-6}
\multicolumn{2}{c}{} & L\'evy & DEM & DE & GA &  \\ \hline
\multirow{3}{*}{GD} & Best & 4.51E-06 & 4.21E-06 & 3.49E-06 & 1.74E-06 & 9.06E-06 \\
 & Median & 5.78E-06 & 7.26E-06 & 7.98E-06 & {\ul \textbf{2.29E-06}} & 1.19E-05 \\
 & Std. & 8.78E-07 & 2.22E-06 & 1.18E-04 & 2.60E-07 & 1.34E-06 \\ \hline
\multirow{3}{*}{S} & Best & 1.56E-05 & 1.50E-05 & 8.51E-06 & 9.38E-06 & 3.94E-05 \\
 & Median & 2.06E-05 & 2.34E-05 & 1.80E-05 & {\ul \textbf{1.53E-05}} & 4.92E-05 \\
 & Std. & 7.18E-06 & 9.07E-06 & 6.76E-06 & 5.71E-06 & 4.19E-06 \\ \hline
\multirow{3}{*}{MS} & Best & 9.13E-03 & 9.23E-03 & 9.00E-03 & 8.92E-03 & 9.06E-03 \\
 & Median & {\ul \textbf{8.96E-03}} & 8.89E-03 & 8.54E-03 & 8.25E-03 & 8.63E-03 \\
 & Std. & 1.01E-04 & 1.98E-04 & 9.93E-04 & 3.16E-04 & 2.18E-04 \\ \hline
\multirow{3}{*}{$\Delta$} & Best & 2.47E-01 & 2.53E-01 & 2.33E-01 & 2.61E-01 & 4.48E-01 \\
 & Median & {\ul \textbf{2.64E-01}} & 2.87E-01 & 2.87E-01 & 2.80E-01 & 4.94E-01 \\
 & Std. & 3.11E-02 & 3.99E-02 & 8.20E-02 & 1.40E-02 & 3.50E-02 \\ \hline
\multirow{3}{*}{IGD} & Best & 2.90E-05 & 2.99E-05 & 3.15E-05 & 2.98E-05 & 3.92E-05 \\
 & Median & {\ul \textbf{3.13E-05}} & 3.50E-05 & 6.03E-05 & 7.54E-05 & 5.01E-05 \\
 & Std. & 2.75E-06 & 8.54E-06 & 2.44E-04 & 3.97E-05 & 1.55E-05 \\ \hline
\multirow{3}{*}{HV} & Best & 2.64E-05 & 2.64E-05 & 2.64E-05 & 2.64E-05 & 2.63E-05 \\
 & Median & 2.64E-05 & 2.63E-05 & 2.63E-05 & {\ul \textbf{2.64E-05}} & 2.63E-05 \\
 & Std. & 9.76E-09 & 2.64E-08 & 2.21E-06 & 1.38E-08 & 1.31E-08 \\ \hline
\end{tabular}
\end{table*}

\begin{table*}[p]
\small\centering
\caption{Numerical results on Hangseng in Experiment II}
\label{tab:res2hangseng}
\begin{tabular}{cccccc}
\hline
\multicolumn{2}{c}{Metric} & LEVY & UNIF & NORM & CONST \\ \hline
\multirow{3}{*}{GD} & Best & 4.95E-06 & 4.57E-06 & 4.24E-06 & 3.49E-06 \\
 & Median & 8.10E-06 & 7.07E-06 & \textbf{7.02E-06} & 7.98E-06 \\
 & Std. & 1.44E-06 & 1.62E-06 & 1.44E-06 & 1.18E-04 \\ \hline
\multirow{3}{*}{S} & Best & 1.43E-05 & 1.43E-05 & 1.37E-05 & 8.51E-06 \\
 & Median & 1.98E-05 & 1.83E-05 & 1.85E-05 & \textbf{1.80E-05} \\
 & Std. & 4.95E-06 & 4.66E-06 & 4.45E-06 & 6.76E-06 \\ \hline
\multirow{3}{*}{MS} & Best & 9.02E-03 & 9.09E-03 & 9.10E-03 & 9.00E-03 \\
 & Median & 8.85E-03 & 8.82E-03 & \textbf{8.86E-03} & 8.54E-03 \\
 & Std. & 1.39E-04 & 2.04E-04 & 1.87E-04 & 9.93E-04 \\ \hline
\multirow{3}{*}{$\Delta$} & Best & 2.48E-01 & 2.47E-01 & 2.47E-01 & 2.33E-01 \\
 & Median & 2.64E-01 & \textbf{2.63E-01} & 2.65E-01 & 2.87E-01 \\
 & Std. & 2.26E-02 & 1.94E-02 & 1.60E-02 & 8.20E-02 \\ \hline
\multirow{3}{*}{IGD} & Best & 2.97E-05 & 2.88E-05 & 2.89E-05 & 3.15E-05 \\
 & Median & 3.40E-05 & 3.40E-05 & \textbf{3.27E-05} & 6.03E-05 \\
 & Std. & 5.92E-06 & 9.76E-06 & 8.27E-06 & 2.44E-04 \\ \hline
\multirow{3}{*}{HV} & Best & 2.64E-05 & 2.64E-05 & 2.64E-05 & 2.64E-05 \\
 & Median & 2.63E-05 & 2.64E-05 & \textbf{2.64E-05} & 2.63E-05 \\
 & Std. & 1.42E-08 & 1.69E-08 & 1.43E-08 & 2.21E-06 \\ \hline
\end{tabular}
\end{table*}

\begin{table*}[p]
\small\centering
\caption{Numerical results on DAX 100 in Experiment I}
\label{tab:res1dax}
\begin{tabular}{ccccccc}
\hline
\multicolumn{2}{c}{\multirow{2}{*}{Metric}} & \multicolumn{4}{c}{MOEA/D} & \multirow{2}{*}{NSGA-II} \\ \cline{3-6}
\multicolumn{2}{c}{} & L\'evy & DEM & DE & GA &  \\ \hline
\multirow{3}{*}{GD} & Best & 5.85E-06 & 7.11E-06 & 7.32E-06 & 1.83E-06 & 5.37E-06 \\
 & Median & 7.98E-06 & 9.53E-06 & 1.65E-05 & {\ul \textbf{2.78E-06}} & 8.04E-06 \\
 & Std. & 1.11E-06 & 1.81E-06 & 9.27E-05 & 6.36E-07 & 1.99E-06 \\ \hline
\multirow{3}{*}{S} & Best & 2.76E-05 & 2.34E-05 & 1.64E-05 & 1.59E-05 & 2.27E-05 \\
 & Median & 3.25E-05 & 3.24E-05 & 2.98E-05 & {\ul \textbf{2.48E-05}} & 4.34E-05 \\
 & Std. & 7.29E-06 & 7.37E-06 & 6.61E-06 & 5.68E-06 & 6.94E-06 \\ \hline
\multirow{3}{*}{MS} & Best & 8.13E-03 & 8.12E-03 & 8.36E-03 & 7.37E-03 & 7.83E-03 \\
 & Median & {\ul \textbf{7.77E-03}} & 7.71E-03 & 7.40E-03 & 6.04E-03 & 7.20E-03 \\
 & Std. & 1.59E-04 & 2.31E-04 & 7.26E-04 & 3.70E-04 & 5.20E-04 \\ \hline
\multirow{3}{*}{$\Delta$} & Best & 3.88E-01 & 4.07E-01 & 3.60E-01 & 4.54E-01 & 5.70E-01 \\
 & Median & {\ul \textbf{4.07E-01}} & 4.49E-01 & 4.36E-01 & 5.81E-01 & 6.68E-01 \\
 & Std. & 2.72E-02 & 3.79E-02 & 7.72E-02 & 3.65E-02 & 5.11E-02 \\ \hline
\multirow{3}{*}{IGD} & Best & 3.28E-05 & 3.62E-05 & 4.40E-05 & 7.20E-05 & 4.14E-05 \\
 & Median & {\ul \textbf{4.16E-05}} & 4.90E-05 & 9.48E-05 & 1.54E-04 & 6.55E-05 \\
 & Std. & 7.87E-06 & 1.60E-05 & 9.39E-05 & 3.75E-05 & 3.34E-05 \\ \hline
\multirow{3}{*}{HV} & Best & 1.91E-05 & 1.90E-05 & 1.90E-05 & 1.91E-05 & 1.90E-05 \\
 & Median & {\ul \textbf{1.90E-05}} & 1.90E-05 & 1.89E-05 & 1.84E-05 & 1.90E-05 \\
 & Std. & 1.16E-08 & 2.34E-08 & 1.08E-06 & 4.22E-07 & 4.44E-07 \\ \hline
\end{tabular}
\end{table*}

\begin{table*}[p]
\small\centering
\caption{Numerical results on DAX 100 in Experiment II}
\label{tab:res2dax}
\begin{tabular}{cccccc}
\hline
\multicolumn{2}{c}{Metric} & LEVY & UNIF & NORM & CONST \\ \hline
\multirow{3}{*}{GD} & Best & 7.23E-06 & 4.76E-06 & 4.88E-06 & 7.32E-06 \\
 & Median & 9.37E-06 & 7.43E-06 & {\ul \textbf{6.85E-06}} & 1.65E-05 \\
 & Std. & 9.63E-07 & 1.39E-06 & 1.47E-06 & 9.27E-05 \\ \hline
\multirow{3}{*}{S} & Best & 2.58E-05 & 2.61E-05 & 2.70E-05 & 1.64E-05 \\
 & Median & 3.28E-05 & 3.09E-05 & 3.19E-05 & {\ul \textbf{2.98E-05}} \\
 & Std. & 3.96E-06 & 3.97E-06 & 5.22E-06 & 6.61E-06 \\ \hline
\multirow{3}{*}{MS} & Best & 8.00E-03 & 8.02E-03 & 8.13E-03 & 8.36E-03 \\
 & Median & 7.75E-03 & 7.76E-03 & {\ul \textbf{7.82E-03}} & 7.40E-03 \\
 & Std. & 1.59E-04 & 1.48E-04 & 1.36E-04 & 7.26E-04 \\ \hline
\multirow{3}{*}{$\Delta$} & Best & 3.85E-01 & 3.94E-01 & 3.89E-01 & 3.60E-01 \\
 & Median & 4.12E-01 & 4.08E-01 & \textbf{4.07E-01} & 4.36E-01 \\
 & Std. & 2.32E-02 & 2.47E-02 & 2.55E-02 & 7.72E-02 \\ \hline
\multirow{3}{*}{IGD} & Best & 3.40E-05 & 3.34E-05 & 3.20E-05 & 4.40E-05 \\
 & Median & 4.39E-05 & 4.07E-05 & {\ul \textbf{3.83E-05}} & 9.48E-05 \\
 & Std. & 1.14E-05 & 7.89E-06 & 5.77E-06 & 9.39E-05 \\ \hline
\multirow{3}{*}{HV} & Best & 1.90E-05 & 1.91E-05 & 1.91E-05 & 1.90E-05 \\
 & Median & 1.90E-05 & 1.90E-05 & \textbf{1.90E-05} & 1.89E-05 \\
 & Std. & 1.19E-08 & 1.28E-08 & 1.50E-08 & 1.08E-06 \\ \hline
\end{tabular}
\end{table*}
\begin{table*}[p]
\small\centering
\caption{Numerical results on FTSE 100 in Experiment I}
\label{tab:res1ftse}
\begin{tabular}{ccccccc}
\hline
\multicolumn{2}{c}{\multirow{2}{*}{Metric}} & \multicolumn{4}{c}{MOEA/D} & \multirow{2}{*}{NSGA-II} \\ \cline{3-6}
\multicolumn{2}{c}{} & L\'evy & DEM & DE & GA &  \\ \hline
\multirow{3}{*}{GD} & Best & 5.39E-06 & 6.32E-06 & 7.01E-06 & 2.84E-06 & 7.38E-06 \\
 & Median & 7.12E-06 & 9.58E-06 & 1.83E-05 & {\ul \textbf{5.05E-06}} & 9.25E-06 \\
 & Std. & 7.30E-07 & 2.41E-06 & 1.55E-04 & 1.60E-06 & 9.51E-07 \\ \hline
\multirow{3}{*}{S} & Best & 1.59E-05 & 1.38E-05 & 9.87E-06 & 1.14E-05 & 2.39E-05 \\
 & Median & 2.09E-05 & 2.01E-05 & {\ul \textbf{1.72E-05}} & 1.97E-05 & 2.99E-05 \\
 & Std. & 3.96E-06 & 4.54E-06 & 3.34E-06 & 5.10E-06 & 2.24E-06 \\ \hline
\multirow{3}{*}{MS} & Best & 5.85E-03 & 5.79E-03 & 5.74E-03 & 5.47E-03 & 5.67E-03 \\
 & Median & {\ul \textbf{5.54E-03}} & 5.40E-03 & 5.15E-03 & 4.91E-03 & 5.45E-03 \\
 & Std. & 1.46E-04 & 1.93E-04 & 5.08E-04 & 4.19E-04 & 1.69E-04 \\ \hline
\multirow{3}{*}{$\Delta$} & Best & 4.06E-01 & 4.25E-01 & 4.21E-01 & 4.27E-01 & 5.47E-01 \\
 & Median & {\ul \textbf{4.33E-01}} & 4.72E-01 & 4.51E-01 & 5.05E-01 & 6.06E-01 \\
 & Std. & 3.38E-02 & 3.06E-02 & 6.30E-02 & 6.88E-02 & 3.32E-02 \\ \hline
\multirow{3}{*}{IGD} & Best & 2.36E-05 & 2.90E-05 & 4.26E-05 & 4.07E-05 & 3.22E-05 \\
 & Median & {\ul \textbf{3.83E-05}} & 5.31E-05 & 9.09E-05 & 8.76E-05 & 4.74E-05 \\
 & Std. & 1.08E-05 & 2.16E-05 & 1.47E-04 & 4.33E-05 & 1.38E-05 \\ \hline
\multirow{3}{*}{HV} & Best & 1.37E-05 & 1.37E-05 & 1.37E-05 & 1.37E-05 & 1.37E-05 \\
 & Median & {\ul \textbf{1.37E-05}} & 1.37E-05 & 1.36E-05 & 1.34E-05 & 1.37E-05 \\
 & Std. & 5.72E-09 & 1.64E-08 & 1.36E-06 & 6.19E-07 & 8.35E-08 \\ \hline
\end{tabular}
\end{table*}

\begin{table*}[p]
\small\centering
\caption{Numerical results on FTSE 100 in Experiment II}
\label{tab:res2ftse}
\begin{tabular}{cccccc}
\hline
\multicolumn{2}{c}{Metric} & LEVY & UNIF & NORM & CONST \\ \hline
\multirow{3}{*}{GD} & Best & 6.56E-06 & 5.58E-06 & 4.83E-06 & 7.01E-06 \\
 & Median & 8.48E-06 & 8.26E-06 & {\ul \textbf{7.26E-06}} & 1.83E-05 \\
 & Std. & 1.12E-06 & 2.24E-06 & 5.79E-05 & 1.55E-04 \\ \hline
\multirow{3}{*}{S} & Best & 1.73E-05 & 1.51E-05 & 1.62E-05 & 9.87E-06 \\
 & Median & 2.06E-05 & 1.75E-05 & 1.87E-05 & \textbf{1.72E-05} \\
 & Std. & 3.94E-06 & 1.56E-06 & 3.02E-06 & 3.34E-06 \\ \hline
\multirow{3}{*}{MS} & Best & 5.82E-03 & 5.48E-03 & 5.69E-03 & 5.74E-03 \\
 & Median & {\ul \textbf{5.47E-03}} & 5.19E-03 & 5.35E-03 & 5.15E-03 \\
 & Std. & 1.54E-04 & 1.29E-04 & 2.07E-04 & 5.08E-04 \\ \hline
\multirow{3}{*}{$\Delta$} & Best & 4.09E-01 & 4.19E-01 & 4.13E-01 & 4.21E-01 \\
 & Median & \textbf{4.35E-01} & 4.45E-01 & 4.36E-01 & 4.51E-01 \\
 & Std. & 2.03E-02 & 1.13E-02 & 2.60E-02 & 6.30E-02 \\ \hline
\multirow{3}{*}{IGD} & Best & 2.62E-05 & 3.95E-05 & 3.02E-05 & 4.26E-05 \\
 & Median & {\ul \textbf{4.29E-05}} & 7.26E-05 & 5.51E-05 & 9.09E-05 \\
 & Std. & 1.27E-05 & 1.66E-05 & 4.61E-05 & 1.47E-04 \\ \hline
\multirow{3}{*}{HV} & Best & 1.37E-05 & 1.37E-05 & 1.37E-05 & 1.37E-05 \\
 & Median & \textbf{1.37E-05} & 1.37E-05 & 1.37E-05 & 1.36E-05 \\
 & Std. & 8.46E-09 & 2.48E-08 & 4.42E-07 & 1.36E-06 \\ \hline
\end{tabular}
\end{table*}
\begin{table*}[p]
\small\centering
\caption{Numerical results on S\&P 100 in Experiment I}
\label{tab:res1sp}
\begin{tabular}{ccccccc}
\hline
\multicolumn{2}{c}{\multirow{2}{*}{Metric}} & \multicolumn{4}{c}{MOEA/D} & \multirow{2}{*}{NSGA-II} \\ \cline{3-6}
\multicolumn{2}{c}{} & L\'evy & DEM & DE & GA &  \\ \hline
\multirow{3}{*}{GD} & Best & 1.05E-05 & 1.13E-05 & 1.27E-05 & 3.20E-06 & 1.01E-05 \\
 & Median & 1.25E-05 & 1.75E-05 & 3.92E-05 & {\ul \textbf{5.02E-06}} & 1.29E-05 \\
 & Std. & 1.64E-06 & 3.45E-06 & 7.64E-05 & 1.86E-06 & 1.42E-06 \\ \hline
\multirow{3}{*}{S} & Best & 2.08E-05 & 2.01E-05 & 1.72E-05 & 1.77E-05 & 2.51E-05 \\
 & Median & 2.51E-05 & 2.76E-05 & \textbf{2.31E-05} & 2.35E-05 & 3.58E-05 \\
 & Std. & 5.10E-06 & 5.26E-06 & 4.78E-06 & 5.06E-06 & 3.68E-06 \\ \hline
\multirow{3}{*}{MS} & Best & 7.72E-03 & 7.86E-03 & 7.60E-03 & 6.87E-03 & 7.29E-03 \\
 & Median & 7.44E-03 & {\ul \textbf{7.55E-03}} & 7.23E-03 & 5.82E-03 & 6.57E-03 \\
 & Std. & 1.47E-04 & 1.61E-04 & 2.53E-04 & 4.35E-04 & 3.52E-04 \\ \hline
\multirow{3}{*}{$\Delta$} & Best & 3.29E-01 & 3.30E-01 & 3.29E-01 & 4.22E-01 & 5.37E-01 \\
 & Median & {\ul \textbf{3.45E-01}} & 3.74E-01 & 3.66E-01 & 5.57E-01 & 6.48E-01 \\
 & Std. & 2.95E-02 & 3.37E-02 & 2.41E-02 & 3.88E-02 & 3.71E-02 \\ \hline
\multirow{3}{*}{IGD} & Best & 3.30E-05 & 3.30E-05 & 3.64E-05 & 4.88E-05 & 4.08E-05 \\
 & Median & \textbf{3.97E-05} & 4.35E-05 & 7.12E-05 & 1.36E-04 & 7.08E-05 \\
 & Std. & 9.23E-06 & 7.65E-06 & 4.16E-05 & 4.13E-05 & 2.21E-05 \\ \hline
\multirow{3}{*}{HV} & Best & 1.87E-05 & 1.87E-05 & 1.87E-05 & 1.87E-05 & 1.87E-05 \\
 & Median & {\ul \textbf{1.87E-05}} & 1.86E-05 & 1.85E-05 & 1.79E-05 & 1.83E-05 \\
 & Std. & 1.51E-08 & 3.19E-08 & 4.79E-07 & 5.03E-07 & 3.13E-07 \\ \hline
\end{tabular}
\end{table*}

\begin{table*}[p]
\small\centering
\caption{Numerical results on S\&P 100 in Experiment II}
\label{tab:res2sp}
\begin{tabular}{cccccc}
\hline
\multicolumn{2}{c}{Metric} & LEVY & UNIF & NORM & CONST \\ \hline
\multirow{3}{*}{GD} & Best & 1.09E-05 & 7.66E-06 & 7.01E-06 & 1.27E-05 \\
 & Median & 1.41E-05 & \textbf{1.01E-05} & 1.04E-05 & 3.92E-05 \\
 & Std. & 1.83E-06 & 2.67E-05 & 2.99E-05 & 7.64E-05 \\ \hline
\multirow{3}{*}{S} & Best & 2.20E-05 & 2.07E-05 & 2.07E-05 & 1.72E-05 \\
 & Median & 2.58E-05 & \textbf{2.29E-05} & 2.42E-05 & 2.31E-05 \\
 & Std. & 4.15E-06 & 2.61E-06 & 4.55E-06 & 4.78E-06 \\ \hline
\multirow{3}{*}{MS} & Best & 7.75E-03 & 7.47E-03 & 7.64E-03 & 7.60E-03 \\
 & Median & 7.32E-03 & 7.20E-03 & \textbf{7.35E-03} & 7.23E-03 \\
 & Std. & 1.60E-04 & 1.33E-04 & 1.65E-04 & 2.53E-04 \\ \hline
\multirow{3}{*}{Delta} & Best & 3.28E-01 & 3.32E-01 & 3.30E-01 & 3.29E-01 \\
 & Median & 3.55E-01 & 3.49E-01 & \textbf{3.49E-01} & 3.66E-01 \\
 & Std. & 2.04E-02 & 1.86E-02 & 3.04E-02 & 2.41E-02 \\ \hline
\multirow{3}{*}{IGD} & Best & 3.35E-05 & 3.62E-05 & 3.12E-05 & 3.64E-05 \\
 & Median & 4.72E-05 & 5.23E-05 & \textbf{4.28E-05} & 7.12E-05 \\
 & Std. & 1.11E-05 & 1.42E-05 & 1.31E-05 & 4.16E-05 \\ \hline
\multirow{3}{*}{HV} & Best & 1.87E-05 & 1.87E-05 & 1.88E-05 & 1.87E-05 \\
 & Median & 1.87E-05 & 1.87E-05 & \textbf{1.87E-05} & 1.85E-05 \\
 & Std. & 1.73E-08 & 1.65E-07 & 1.78E-07 & 4.79E-07 \\ \hline
\end{tabular}
\end{table*}
\begin{table*}[p]
\small\centering
\caption{Numerical results on Nikkei in Experiment I}
\label{tab:res1nikkei}
\begin{tabular}{ccccccc}
\hline
\multicolumn{2}{c}{\multirow{2}{*}{Metric}} & \multicolumn{4}{c}{MOEA/D} & \multirow{2}{*}{NSGA-II} \\ \cline{3-6}
\multicolumn{2}{c}{} & L\'evy & DEM & DE & GA &  \\ \hline
\multirow{3}{*}{GD} & Best & 5.45E-06 & 5.10E-06 & 5.93E-05 & 9.95E-06 & 2.54E-06 \\
 & Median & 7.26E-06 & 8.06E-06 & 1.45E-04 & 3.31E-05 & {\ul \textbf{4.24E-06}} \\
 & Std. & 1.16E-06 & 1.66E-04 & 7.09E-05 & 2.08E-04 & 2.01E-06 \\ \hline
\multirow{3}{*}{S} & Best & 1.28E-05 & 1.28E-05 & 5.99E-06 & 0.00E+00 & 1.05E-05 \\
 & Median & 1.76E-05 & 2.08E-05 & {\ul \textbf{1.15E-05}} & 1.92E-05 & 1.45E-05 \\
 & Std. & 2.88E-06 & 5.77E-06 & 4.53E-06 & 9.72E-06 & 1.84E-06 \\ \hline
\multirow{3}{*}{MS} & Best & 4.09E-03 & 4.23E-03 & 4.29E-03 & 2.63E-03 & 3.36E-03 \\
 & Median & 3.93E-03 & \textbf{3.94E-03} & 2.96E-03 & 2.20E-03 & 2.88E-03 \\
 & Std. & 1.38E-04 & 5.39E-04 & 5.48E-04 & 4.65E-04 & 2.54E-04 \\ \hline
\multirow{3}{*}{$\Delta$} & Best & 3.94E-01 & 3.99E-01 & 3.17E-01 & 8.40E-01 & 6.09E-01 \\
 & Median & {\ul \textbf{4.34E-01}} & 4.81E-01 & 5.58E-01 & 9.34E-01 & 6.81E-01 \\
 & Std. & 3.89E-02 & 9.05E-02 & 1.00E-01 & 3.54E-02 & 2.95E-02 \\ \hline
\multirow{3}{*}{IGD} & Best & 1.77E-05 & 1.90E-05 & 7.90E-05 & 1.77E-04 & 4.64E-05 \\
 & Median & \textbf{2.39E-05} & 2.73E-05 & 2.23E-04 & 2.41E-04 & 9.69E-05 \\
 & Std. & 1.15E-05 & 4.09E-04 & 6.95E-05 & 5.29E-04 & 3.67E-05 \\ \hline
\multirow{3}{*}{HV} & Best & 8.31E-06 & 8.29E-06 & 7.96E-06 & 7.87E-06 & 8.19E-06 \\
 & Median & {\ul \textbf{8.29E-06}} & 8.26E-06 & 7.23E-06 & 7.54E-06 & 7.94E-06 \\
 & Std. & 1.59E-08 & 9.52E-07 & 3.43E-07 & 1.20E-06 & 1.08E-07 \\ \hline
\end{tabular}
\end{table*}

\begin{table*}[p]
\small\centering
\caption{Numerical results on Nikkei in Experiment II}
\label{tab:res2nikkei}
\begin{tabular}{cccccc}
\hline
\multicolumn{2}{c}{Metric} & LEVY & UNIF & NORM & CONST \\ \hline
\multirow{3}{*}{GD} & Best & 5.92E-06 & 5.45E-06 & 5.38E-06 & 5.93E-05 \\
 & Median & {\ul \textbf{7.71E-06}} & 4.25E-05 & 4.10E-05 & 1.45E-04 \\
 & Std. & 1.42E-06 & 1.93E-04 & 2.65E-05 & 7.09E-05 \\ \hline
\multirow{3}{*}{S} & Best & 9.54E-06 & 1.24E-05 & 8.75E-06 & 5.99E-06 \\
 & Median & 1.66E-05 & 1.54E-05 & 1.50E-05 & {\ul \textbf{1.15E-05}} \\
 & Std. & 5.35E-06 & 2.62E-06 & 3.05E-06 & 4.53E-06 \\ \hline
\multirow{3}{*}{MS} & Best & 4.14E-03 & 4.08E-03 & 4.07E-03 & 4.29E-03 \\
 & Median & {\ul \textbf{3.90E-03}} & 3.68E-03 & 3.70E-03 & 2.96E-03 \\
 & Std. & 1.57E-04 & 5.59E-04 & 3.12E-04 & 5.48E-04 \\ \hline
\multirow{3}{*}{$\Delta$} & Best & 3.89E-01 & 3.84E-01 & 3.76E-01 & 3.17E-01 \\
 & Median & {\ul \textbf{4.33E-01}} & 4.44E-01 & 4.55E-01 & 5.58E-01 \\
 & Std. & 5.61E-02 & 9.82E-02 & 6.73E-02 & 1.00E-01 \\ \hline
\multirow{3}{*}{IGD} & Best & 1.84E-05 & 1.83E-05 & 1.95E-05 & 7.90E-05 \\
 & Median & {\ul \textbf{2.72E-05}} & 4.16E-05 & 4.85E-05 & 2.23E-04 \\
 & Std. & 1.68E-05 & 4.51E-04 & 2.98E-05 & 6.95E-05 \\ \hline
\multirow{3}{*}{HV} & Best & 8.31E-06 & 8.29E-06 & 8.28E-06 & 7.96E-06 \\
 & Median & {\ul \textbf{8.29E-06}} & 8.11E-06 & 8.10E-06 & 7.23E-06 \\
 & Std. & 2.80E-08 & 9.83E-07 & 1.34E-07 & 3.43E-07 \\ \hline
\end{tabular}
\end{table*}

\section{Discussion} \label{sec:discussion}
The numerical results show a good performance of the proposed MOEA/D-L\'evy. It is easy to realize that those long trajectories caused by LF can enhance the global search capability of algorithms by comparing the results of LEVY and NORM in Experiment II, as the main difference between a heavy-distribution and standard normal distribution is the occasional generation of large numbers. In this section, we show further insights on how long trajectories contribute to this observed improvement.

To record long trajectories, we compute the Euclidean distance between a repaired offspring $\boldsymbol{y}$ and its parent $\boldsymbol{x^i}$ in the mutation methods based on LEVY, UNIF, NORM, and CONST, using Nikkei dataset. This distance is the length of the trial vector. In addition, we record the frequency that this offspring is successfully updated. Based on the parameter settings in \textbf{Section~\ref{sec:experimental-method}}, the possible frequency for one offspring is 0, 1 or 2. \figref{fig:trialpopexp2gen1a},~\ref{fig:trialpopexp2gen3a} and~\ref{fig:trialpopexp2gen10a} show the frequency of trajectories in different length, at 1st, 3rd and 10th generation in one certain run, respectively. \figref{fig:trialpopexp2gen1b},~\ref{fig:trialpopexp2gen3b} and~\ref{fig:trialpopexp2gen10b} illustrate the corresponding populations at each generation. \figref{fig:5alg10thpoponnikkei} shows a population on objective space at 10th generation in Experiment I. \figref{fig:trajbygen} shows the frequency of successfully updated long trajectories (i.e., the length is larger than 0.2) by generations in MOEA/D-L\'evy in the same run.

It is interesting to notice that there are some long trial vectors successfully updated in the beginning phase of LEVY, while the same observation does not occur in UNIF, NORM, and CONST. These long trajectories usually represent a global search. As a result, LEVY achieves early in the optimization a solution set that is widely spread across the objective space (i.e., show in \figref{fig:trialpopexp2gen1b} and~\figref{fig:trialpopexp2gen3b}). In~\figref{fig:5alg10thpoponnikkei} and~\ref{fig:trialpopexp2gen10b}, methods based on LF form a relatively better front than other methods. \figref{fig:popexp1a} to~\figref{fig:popexp1f} illustrates the population of five algorithms in Experiment I at different generations in a certain run. In the beginning phase of the optimization, MOEA/D-L\'evy forms multiple ``sub-sets" while the other methods hold only one front. This may indicate that methods based on LF can search multiple areas at the same time, while other mutations can only deal with one.

To explain this behavior, consider that, in PO, trial vectors or trajectories represent the re-allocation of capital. The heavy-tail distribution holds a large probability to generate small values, and a small probability to generate large values.
\figref{fig:poexplan} presents an example of this procedure. The variable (asset) that receives a large number during mutation will get more capital allocation. What is more, to satisfy the constrain that the summation of variables equals 1 in~\eqref{eq:m-vconstraints}, the repair steps will implement proper scaling, and thus the other variables (assets) will be reduced. Therefore, the re-allocation of capital will be centralized on some assets rather than equally distributed across all assets. If these assets hold relatively high returns, the algorithm can find portfolio candidates with high returns early in the optimization. As the initial population is centralized at the low-risk area in objective space because of the uniform initialization, this mutated candidate has a large probability to be successfully updated. Following this procedure, the solution set will be distributed in multiple areas on the search space. Then, the algorithm will mainly update solutions by local search but in multiple search areas, as the accepted long trials decrease by generation in~\figref{fig:trajbygen}. Such different search patterns may lead to the observed improvement compared with other methods.

\section{Conclusions} \label{sec:conclusion}
In this study, we have proposed a novel method to solve the MOOP formulation of PO by injecting LF into MOEA/D as a mutation method. The simulation results have indicated that the proposed method holds better performance in most cases, compared with MOEA/D-DEM, MOEA/D-DE, MOEA/D-GA, and NSGA-II. In addition, we have compared the LF mutation with three other probability-based mutations. On the largest benchmark, Nikkei, LEVY shows better performance. Further insight into the evolutionary process indicated that algorithms based on LF mutation promote global search at the beginning and search multiple areas of the objective space at the same time. This search strategy contributes to improvement.

In the future, we want to extend the proposed method to a more constrained PO to study how LF work with constraints. In~\figref{fig:trajbygen}, few long trajectories get successfully updated with the generation increased. Thus, we may design an adaptive strategy to reduce global search but enhance local search in the middle and ending period of optimization. It is also interesting to apply this method on other resource allocation problems.

\section*{Acknowledgements}
We would like to thank Prof.~Hitoshi Kanoh for several contributions to this manuscript.
This research did not receive any specific grant from funding agencies in the public, commercial, or not-for-profit sectors.

\bibliographystyle{ws-adsaa}       
\bibliography{refs}             

\end{document}